\documentclass[accepted,article,table,xcdraw]{uai2021} 

\usepackage[american]{babel}

\usepackage[square,authoryear]{natbib} 
\usepackage{mathtools} 
\usepackage{booktabs} 
\usepackage{tikz} 
\usepackage{microtype}
\usepackage{graphicx}
\usepackage{subfig}
\usepackage{caption}
\usepackage{booktabs} 
\usepackage{multirow}
\usepackage{amsmath,amssymb}
\usepackage{setspace}
\usepackage{comment}
\usepackage{bm}
\usepackage{dsfont}
\usepackage{authblk}

\DeclareMathOperator*{\argmin}{argmin}
\usepackage{amsthm}
\theoremstyle{definition}
\newtheorem{defn}{Definition}
\newtheorem{theorem}{Theorem}
\newtheorem{remark}{Remark}
\usepackage{enumitem}
\newtheorem{prop}{Proposition}
\newtheorem*{prop*}{Proposition}
\usepackage{algorithm}
\usepackage{algorithmic}
\newtheorem*{theorem*}{Theorem}
\usepackage{sidecap}
\usepackage[toc,page]{appendix}

\usepackage{hyperref}

\newcommand{\inp}{\bm{x}_i}

\newcommand{\cfinp}{\bm{x}}

\newcommand\factorspace{\mathcal{C}}

\newcommand\factor{c}
\newcommand\factorset{C}

\newcommand\toposortfn{\mathsf{topological\_sort}}

\newcommand\modelfn{f}
\newcommand{\indep}{\rotatebox[origin=c]{90}{$\models$}}

\usepackage{tikz}
\usetikzlibrary{cd,arrows,calc,shapes,positioning}
\tikzstyle{obs} = [circle,fill=white,draw=black,inner sep=1pt,minimum size=20pt,font=\fontsize{10}{10}\selectfont,node distance=1,thick]
\tikzstyle{latent} = [obs,dotted]
\tikzstyle{plate} = [draw, rectangle, minimum size=20pt]
\newcommand{\edge}[3][]{ %
  \foreach \x in {#2} { %
    \foreach \y in {#3} { %
      \path (\x) edge [->, >={triangle 45}, #1,thick] (\y) ;%
    } ;
  } ;
}

\newcommand*\mytabalign{%
    \edef\sk@align{\ifodd\c@page l\else r\fi}
    \makebox[\textwidth][\sk@align]}
\makeatother

\makeatletter 
    \renewcommand\AB@affilsepx{  \protect\Affilfont} 
\makeatother

\title{Local Explanations via Necessity and Sufficiency:\\ 
Unifying Theory and Practice}

\author[*1]{\href{mailto:David S. Watson <david.watson@ucl.ac.uk>?Subject=Your UAI 2021 paper}{David S. Watson}}
\author[*2,3]{\href{mailto:Limor Gultchin <limor.gultchin@gmail.com>?Subject=Your UAI 2021 paper}{Limor Gultchin}}
\author[4]{Ankur Taly}
\author[5,3]{Luciano Floridi}
\affil[*]{Equal contribution}
\affil[1]{Department of Statistical Science, University College London, London, UK \par}
\affil[2]{Department of Computer Science, University of Oxford, Oxford, UK \par}
\affil[3]{The Alan Turing Institute, London, UK}
\affil[4]{Google Inc., Mountain View, USA \par}
\affil[5]{Oxford Internet Institute, University of Oxford, Oxford, UK}

\begin{document}
\maketitle

\begin{abstract}
Necessity and sufficiency are the building blocks of all successful explanations. Yet despite their importance, these notions have been conceptually underdeveloped and inconsistently applied in explainable artificial intelligence (XAI), a fast-growing research area that is so far lacking in firm theoretical foundations. Building on work in logic, probability, and causality, we establish the central role of necessity and sufficiency in XAI, unifying seemingly disparate methods in a single formal framework. We provide a sound and complete algorithm for computing explanatory factors with respect to a given context, and demonstrate its flexibility and competitive performance against state of the art alternatives on various tasks.
\end{abstract}

\section{Introduction}\label{sec:intro}
Machine learning algorithms are increasingly used in a variety of high-stakes domains, from credit scoring to medical diagnosis. However, many such methods are \emph{opaque}, in that humans cannot understand the reasoning behind particular predictions. Post-hoc, model-agnostic local explanation tools (e.g., feature attributions, rule lists, and counterfactuals) are at the forefront of a fast-growing area of research variously referred to as \emph{interpretable machine learning} or \emph{explainable artificial intelligence} (XAI).

\begin{figure}[!ht]
    \centering
    \includegraphics[scale=0.35]{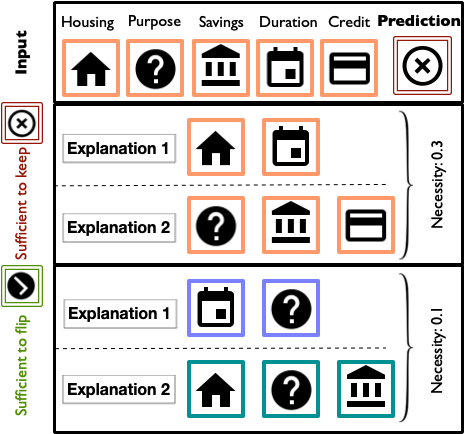}
    \caption{
    We describe minimal sufficient factors (here, sets of features) for a given input (top row), with the aim of preserving or flipping the original prediction. We report a sufficiency score for each set and a cumulative necessity score for all sets, indicating the proportion of paths towards the outcome that are covered by the explanation. Feature colors indicate source of feature values (input or reference).}
    \vspace{-3.7ex}
    \label{fig:general_approach}
\end{figure}

Many authors have pointed out the inconsistencies between popular XAI tools, raising questions as to which method is more reliable in particular cases \citep{mothilal2020_unif, Ramon2020, fernandez_loria_2020}. Theoretical foundations have proven elusive in this area, perhaps due to the perceived subjectivity inherent to notions such as ``intelligible” and ``relevant” \citep{Watson_exp_game}. Practitioners often seek refuge in the axiomatic guarantees of Shapley values, which have become the de facto standard in many XAI applications, due in no small part to their attractive theoretical properties \citep{bhatt_deployment}. However, ambiguities regarding the underlying assumptions of the method \citep{kumar2020} and the recent proliferation of mutually incompatible implementations \citep{Sundararajan2019, taly2020} have complicated this picture. Despite the abundance of alternative XAI tools \citep{Molnar2019}, a dearth of theory persists. This has led some to conclude that the goals of XAI are underspecified \citep{Lipton2018}, and even that post-hoc methods do more harm than good \citep{Rudin2019}.


We argue that this lacuna at the heart of XAI should be filled by a return to fundamentals -- specifically, to \emph{necessity} and \emph{sufficiency}. As the building blocks of all successful explanations, these dual concepts deserve a privileged position in the theory and practice of XAI. Following a review of related work (Sect.~\ref{sec:nec_suf}), we operationalize this insight with a unified framework (Sect. \ref{sec:framework}) that reveals unexpected affinities between various XAI tools and probabilities of causation (Sect. \ref{sec:related}). We proceed to implement a novel procedure for computing model explanations that improves upon the state of the art in various quantitative and qualitative comparisons (Sect.~\ref{sec:experiments}). Following a brief discussion (Sect.~\ref{sec:discussion}), we conclude with a summary and directions for future work (Sect.~\ref{sec:conclusion}).



We make three main contributions. (1) We present a formal framework for XAI that unifies several popular approaches, including feature attributions, rule lists, and counterfactuals. (2) We introduce novel measures of necessity and sufficiency that can be computed for any feature subset. The method enables users to incorporate domain knowledge, search various subspaces, and select a utility-maximizing explanation. (3) We present a sound and complete algorithm for identifying explanatory factors, and illustrate its performance on a range of tasks. 

\section{Necessity and Sufficiency}\label{sec:nec_suf}

Necessity and sufficiency have a long philosophical tradition \citep{Mackie1965, Lewis1973, halpern2005causes2}, spanning logical, probabilistic, and causal variants. In propositional logic, we say that $x$ is a sufficient condition for $y$ iff $x \rightarrow y$, and $x$ is a necessary condition for $y$ iff $y \rightarrow x$. So stated, necessity and sufficiency are logically \emph{converse}. However, by the law of contraposition, both definitions admit alternative formulations, whereby sufficiency may be rewritten as $\neg y \rightarrow \neg x$ and necessity as $\neg x \rightarrow \neg y$. By pairing the original definition of sufficiency with the latter definition of necessity (and vice versa), we find that the two concepts are also logically \emph{inverse}. 

These formulae suggest probabilistic relaxations, measuring $x$'s sufficiency for $y$ by $P(y|x)$ and $x$'s necessity for $y$ by $P(x|y)$. Because there is no probabilistic law of contraposition, these quantities are generally uninformative w.r.t. $P(\neg x|\neg y)$ and $P(\neg y|\neg x)$, which may be of independent interest. Thus, while necessity is both the converse and inverse of sufficiency in propositional logic, the two formulations come apart in probability calculus. We revisit the distinction between probabilistic conversion and inversion in Rmk.~\ref{rmk:pr} and Sect.~\ref{sec:related}. 

These definitions struggle to track our intuitions when we consider causal explanations \citep{Pearl2000, tian2000probabilities}. It may make sense to say in logic that if $x$ is a necessary condition for $y$, then $y$ is a sufficient condition for $x$; it does not follow that if $x$ is a necessary \emph{cause} of $y$, then $y$ is a sufficient \emph{cause} of $x$. We may amend both concepts using \emph{counterfactual probabilities} -- e.g., the probability that Alice would still have a headache if she had not taken an aspirin, given that she does not have a headache and did take an aspirin. Let $P(y_x|x', y')$ denote such a quantity, to be read as ``the probability that $Y$ would equal $y$ under an intervention that sets $X$ to $x$, given that we observe $X = x'$ and $Y = y'$.'' Then, according to \citet[Ch.~9]{Pearl2000}, the probability that $x$ is a sufficient cause of $y$ is given by $\texttt{suf}(x, y) := P(y_x|x', y')$, and the probability that $x$ is a necessary cause of $y$ is given by $\texttt{nec}(x, y) := P(y'_{x'}|x,y).$ 

Analysis becomes more difficult in higher dimensions, where variables may interact to block or unblock causal pathways. \citet{Vanderweele2008} analyze sufficient causal interactions in the potential outcomes framework, refining notions of synergism without monotonicity constraints. In a subsequent paper, \citet{Vanderweele2012} study the irreducibility and singularity of interactions in sufficient-component cause models. \citet{halpern2016actual} devotes an entire monograph to the subject, providing various criteria to distinguish between subtly different notions of “actual causality”, as well as “but-for” (similar to necessary) and sufficient causes. These authors generally limit their analyses to Boolean systems with convenient structural properties, e.g. conditional ignorability and the stable unit treatment value assumption \citep{Imbens2015}. Operationalizing their theories in a practical method without such restrictions is one of our primary contributions.  

Necessity and sufficiency have begun to receive explicit attention in the XAI literature. \citet{Ribeiro2018} propose a bandit procedure for identifying a minimal set of Boolean conditions that entails a predictive outcome (more on this in Sect.~\ref{sec:related}). \citet{pertinent_negatives} propose an autoencoder for learning pertinent negatives and positives, i.e. features whose presence or absence is decisive for a given label, while \citet{zhang_2018} develop a technique for generating symbolic corrections to alter model outputs. Both methods are optimized for neural networks, unlike the model-agnostic approach we develop here. 

Another strand of research in this area is rooted in logic programming. Several authors have sought to reframe XAI as either a SAT \citep{ignatiev2019, Narodytska2019} or a set cover problem \citep{lakkaraju2019, grover2019}, typically deriving approximate solutions on a prespecified subspace to ensure computability in polynomial time. We adopt a different strategy that prioritizes completeness over efficiency, an approach we show to be feasible in moderate dimensions (see Sect.~\ref{sec:discussion} for a discussion). 

\citet{mothilal2020_unif} build on \citet{halpern2016actual}'s definitions of necessity and sufficiency to critique popular XAI tools, proposing a new feature attribution measure with some purported advantages. Their method relies on the strong assumption that predictors are mutually independent. \citet{galhotra2021} adapt \citet{Pearl2000}'s probabilities of causation for XAI under a more inclusive range of data generating processes. They derive analytic bounds on multidimensional extensions of \texttt{nec} and \texttt{suf}, as well as an algorithm for point identification when graphical structure permits. Oddly, they claim that non-causal applications of necessity and sufficiency are somehow “incorrect and misleading” (p. 2), a normative judgment that is inconsistent with many common uses of these concepts. 

Rather than insisting on any particular interpretation of necessity and sufficiency, we propose a general framework that admits logical, probabilistic, and causal interpretations as special cases. Whereas previous works evaluate individual predictors, we focus on feature \emph{subsets}, allowing us to detect and quantify interaction effects. Our formal results clarify the relationship between existing XAI methods and probabilities of causation, while our empirical results demonstrate their applicability to a wide array of tasks and datasets.

\section{A Unifying Framework}\label{sec:framework}
We propose a unifying framework that highlights the role of necessity and sufficiency in XAI. Its constituent elements are described below.
\vspace{-2ex}
\paragraph{Target function.} Post-hoc explainability methods assume access to a target function $f: \mathcal{X} \mapsto \mathcal{Y}$, i.e. the model whose prediction(s) we seek to explain. For simplicity, we restrict attention to the binary setting, with $Y \in \{0, 1\}$. Multi-class extensions are straightforward, while continuous outcomes may be accommodated via discretization. Though this inevitably involves some information loss, we follow authors in the contrastivist tradition in arguing that, even for continuous outcomes, explanations always involve a juxtaposition (perhaps implicit) of ``fact and foil'' \citep{lipton_1990}. For instance, a loan applicant is probably less interested in knowing why her credit score is precisely $y$ than she is in discovering why it is below some threshold (say, 700). Of course, binary outcomes can approximate continuous values with arbitrary precision over repeated trials.


\paragraph{Context.} The context $\mathcal{D}$ is a probability distribution over which we quantify sufficiency and necessity.
Contexts may be constructed in various ways but always consist of at least some input (point or space) and reference (point or space). For instance, we may want to compare $\bm{x}_i$ with all other samples, or else just those perturbed along one or two axes, perhaps based on some conditioning event(s). 

In addition to predictors and outcomes, we optionally include information exogenous to $f$. For instance, if any events were conditioned upon to generate a given reference sample, this information may be recorded among a set of auxiliary variables $\bm{W}$. 
Other examples of potential auxiliaries include metadata or engineered features such as those learned via neural embeddings. This augmentation allows us to evaluate the necessity and sufficiency of factors beyond those found in $\bm{X}$. Contextual data take the form $\bm{Z} = (\bm{X}, \bm{W}) \sim \mathcal{D}$. The distribution may or may not encode dependencies between (elements of) $\bm{X}$ and (elements of) $\bm{W}$. We extend the target function to augmented inputs by defining $f(\bm{z}) := f(\bm{x})$.



\paragraph{Factors.} 
Factors pick out the properties whose necessity and sufficiency we wish to quantify. Formally, a factor $c: \mathcal{Z} \mapsto \{0, 1\}$ indicates whether its argument satisfies some criteria with respect to predictors or auxiliaries. For instance, if $\bm{x}$ is an input to a credit lending model, and $\bm{w}$ contains information about the subspace from which data were sampled, then a factor could be $c(\bm{z}) = \mathds{1}[\bm{x}[\mathsf{gender} = \text{``female''}] \land \bm{w}[do(\mathsf{income} > \$50\text{k})]]$, i.e. checking if $\bm{z}$ is female and drawn from a context in which an intervention fixes income at greater than \$50k. We use the term ``factor'' as opposed to ``condition'' or ``cause'' to suggest an inclusive set of criteria that may apply to predictors $\bm{x}$ and/or auxiliaries $\bm{w}$. Such criteria are always observational w.r.t. $\bm{z}$ but may be interventional or counterfactual w.r.t. $\bm{x}$. We assume a finite space of factors $\factorspace$.
\vspace*{-1ex}
\paragraph{Partial order.} When multiple factors pass a given necessity or sufficiency threshold, users will tend to prefer some over others. For instance, factors with fewer conditions are often preferable to those with more, all else being equal; factors that change a variable by one unit as opposed to two are preferable, and so on. Rather than formalize this preference in terms of a distance metric, which unnecessarily constrains the solution space, we treat the partial ordering as primitive and require only that it be complete and transitive. This covers not just distance-based measures but also more idiosyncratic orderings that are unique to individual agents. Ordinal preferences may be represented by cardinal utility functions under reasonable assumptions (see, e.g., \citep{VonNeumann1944}).

We are now ready to formally specify our framework.
\begin{defn}[Basis]\label{def:framework}
    A \emph{basis} for computing necessary and sufficient factors for model predictions is a tuple $\mathcal{B} = \langle f, \mathcal{D}, \factorspace, \preceq \rangle$, where $f$ is a target function, $\mathcal{D}$ is a context, $\factorspace$ is a set of factors, and $\preceq$ is a partial ordering on $\factorspace$. 
\end{defn}

\subsection{Explanatory Measures}
For some fixed basis $\mathcal{B} = \langle f, \mathcal{D}, \factorspace, \preceq \rangle$, we define the following measures of sufficiency and necessity, with probability taken over $\mathcal{D}$. 


\begin{defn}[Probability of Sufficiency]\label{def:probsuff}
    The probability that $c$ is a sufficient factor for outcome $y$ is given by: 
    \begin{align*}
        PS(c, y) := P(f(\bm{z}) = y~|~c(\bm{z}) = 1). 
    \end{align*} 
    The probability that factor set $\factorset = \{c_1, \dots, c_k\}$ is sufficient for $y$ is given by:
    \begin{align*}
        PS(\factorset, y) := P(f(\bm{z}) = y~|~\sum_{i=1}^k c_i(\bm{z}) \geq 1).
    \end{align*}
\end{defn}

\begin{defn}[Probability of Necessity]\label{def:probnecc}
    The probability that $c$ is a necessary factor for outcome $y$ is given by:
    \begin{align*}
        PN(c, y) := P(c(\bm{z}) = 1~|~f(\bm{z}) = y). 
    \end{align*}
    The probability that factor set $\factorset = \{c_1, \dots, c_k\}$ is necessary for $y$ is given by:
    \begin{align*}
        PN(\factorset, y) := P(\sum_{i=1}^k c_i(\bm{z}) \geq 1~|~f(\bm{z}) = y).
    \end{align*}
\end{defn}

\begin{remark}\label{rmk:pr}
    These probabilities can be likened to the ``precision'' (positive predictive value) and ``recall'' (true positive rate) of a (hypothetical) classifier that predicts whether $f(\bm{z}) = y$ based on whether $c(\bm{z}) = 1$. By examining the confusion matrix of this classifier, one can define other related quantities, e.g. the true negative rate $P(c(\bm{z}) = 0|f(\bm{z}) \neq y)$ and the negative predictive value $P(f(\bm{z}) \neq y|c(\bm{z}) = 0)$, which are contrapositive transformations of our proposed measures. We can recover these values exactly via $PS(1 - c, 1 - y)$ and $PN(1 - c, 1 - y)$, respectively. When necessity and sufficiency are defined as probabilistic inversions (rather than conversions), such transformations are impossible.  
\end{remark}


\subsection{Minimal Sufficient Factors}\label{sec:lens}
We introduce Local Explanations via Necessity and Sufficiency (LENS), a procedure for computing explanatory factors with respect to a given basis $\mathcal{B}$ and threshold parameter $\tau$ (see Alg.~\ref{alg:main}). First, we calculate a factor's probability of sufficiency (see \textsf{probSuff}) by drawing $n$ samples from $\mathcal{D}$ and taking the maximum likelihood estimate $\hat{PS}(c, y)$. Next, we sort the space of factors w.r.t. $\preceq$ in search of those that are $\tau$-minimal.

\begin{defn}[$\tau$-minimality]\label{def:minimality}
    We say that $c$ is $\tau$-minimal iff (i) $PS(c, y) \geq \tau$ and (ii) there exists no factor $c'$ such that  $PS(c', y) \geq \tau$ and $c' \prec c$.
\end{defn}


Since a factor is necessary to the extent that it covers all possible pathways towards a given outcome, our next step is to span the $\tau$-minimal factors and compute their cumulative $PN$ (see \textsf{probNec}). As a minimal factor $\factor$ stands for all $\factor'$ such that $\factor \preceq \factor'$, in reporting probability of necessity, we expand $\factorset$ to its upward closure. 

Thms.~\ref{thm:complete} and \ref{thm:ump} state that this procedure is \emph{optimal} in a sense that depends on whether we assume access to oracle or sample estimates of $PS$ (see Appendix~\ref{sec:app_proofs} for all proofs).


\begin{theorem}\label{thm:complete}
    With oracle estimates $PS(c, y)$ for all $c \in \mathcal{C}$, Alg.~\ref{alg:main} is sound and complete. That is, for any $\factorset$ returned by Alg.~\ref{alg:main} and all $\factor \in \factorspace$, $\factor$ is $\tau$-minimal iff $\factor \in \factorset$. 
\end{theorem}

Population proportions may be obtained if data fully saturate the space $\mathcal{D}$, a plausible prospect for categorical variables of low to moderate dimensionality. Otherwise, proportions will need to be estimated. 

\begin{theorem}\label{thm:ump}
    With sample estimates $\hat{PS}(c, y)$ for all $c \in \mathcal{C}$, Alg.~\ref{alg:main} is uniformly most powerful. That is, Alg.~\ref{alg:main} identifies the most $\tau$-minimal factors of any method with fixed type I error $\alpha$. 
\end{theorem}

Multiple testing adjustments can easily be accommodated, in which case modified optimality criteria apply \citep{Storey2007}.

\begin{remark}\label{rmk:tau}
    We take it that the main quantity of interest in most applications is sufficiency, be it for the original or alternative outcome, and therefore define $\tau$-minimality w.r.t. sufficient (rather than necessary) factors. However, necessity serves an important role in tuning $\tau$, as there is an inherent trade-off between the parameters. More factors are excluded at higher values of $\tau$, thereby inducing lower cumulative $PN$; more factors are included at lower values of $\tau$, thereby inducing higher cumulative $PN$. See Appendix \ref{sec:app_method}.
\end{remark}
\vspace{-2ex}
\begin{algorithm}[!ht]
  \caption{LENS}\label{alg:main}
 \begin{algorithmic}[1]
  \STATE {\bfseries Input:} {$\mathcal{B} = \langle \modelfn, \mathcal{D}, \factorspace, \preceq \rangle, \tau$}
  \STATE {\bfseries Output:} {Factor set $C$, $(\forall c \in C) ~PS(c, y), PN(C, y)$}
\item[]
  \STATE Sample $\hat{D} = \{\bm{z}_i\}_{i=1}^n \sim \mathcal{D}$
  \item[]
  \STATE {\bfseries function} \textsf{probSuff}{($\factor$, $y$)}
      \begin{ALC@g}
      \STATE n(c\&y) = $\sum_{i=1}^n \mathds{1}[c(\bm{z}_i) = 1 \land f(\bm{z}_i) = y]$
        \STATE n(c) = $\sum_{i=1}^n c(\bm{z}_i)$
        \STATE {\bfseries return} n(c\&y)~/~n(c)
      \end{ALC@g}
\item[]
  \STATE {\bfseries function} \textsf{probNec}{($\factorset$, $y$, upward\_closure\_flag)}
    \begin{ALC@g}
      \IF{upward\_closure\_flag}
          \STATE $\factorset  =  \{\factor~|~\factor \in \factorspace \land \exists~\factor' \in \factorset: \factor' \preceq \factor\}$
      \ENDIF 
      \STATE n(C\&y) = $\sum_{i=1}^n \mathds{1}[\sum_{j=1}^k c_j(\bm{z}_i) \geq 1 \land f(\bm{z}_i) = y]$
      \STATE n(y) = $\sum_{i=1}^n \mathds{1}[f(\bm{z}_i) = y]$ 
      \STATE {\bfseries return} n(C\&y)~/~n(y)
    \end{ALC@g}
\item[]
  \STATE {\bfseries function} \textsf{minimalSuffFactors}{($y$, $\tau$, sample\_flag, $\alpha$)}
    \begin{ALC@g}
      \STATE sorted\_factors = $\toposortfn(\factorspace, \preceq)$
      \STATE cands = []
      \FOR {$\factor$ in sorted\_factors}
        \IF {$\exists (\factor', \_) \in\mbox{ cands }: \factor'\preceq\factor$}
          \STATE {\bfseries continue}
        \ENDIF
        \STATE ps = \textsf{probSuff}($\factor$, $y$)
         \IF {sample\_flag}
           \STATE{p = \textsf{binom.test}(n(c\&y), n(c), $\tau$, alt = $>$)}
             \IF{p $\leq \alpha$}
               \STATE cands.append($\factor$, ps)
             \ENDIF
         \ELSIF {ps $\geq \tau$}
             \STATE cands.append($\factor$, ps)
         \ENDIF
    \ENDFOR
    \STATE cum\_pn = \textsf{probNec}{($\{\factor~
      |~(\factor, \_)\in\mbox{cands}\}, y$, \sc{True})}
    \STATE {\bfseries return} cands, cum\_pn
    \end{ALC@g}
 \end{algorithmic}
\end{algorithm}

\vspace{-1ex}

\section{Encoding Existing Measures}\label{sec:related}

Explanatory measures can be shown to play a central role in many seemingly unrelated XAI tools, albeit under different assumptions about the basis tuple $\mathcal{B}$. In this section, we relate our framework to a number of existing methods.

\paragraph{Feature attributions.}\label{sec:attributions} 
Several popular feature attribution algorithms are based on Shapley values \citep{Shapley1953}, which decompose the predictions of any target function as a sum of weights over $d$ input features:
\begin{equation} \label{eq:additive}
    f(\inp) = \phi_0 + \sum_{j=1}^{d}\phi_j,
\end{equation}
where $\phi_0$ represents a baseline expectation and $\phi_j$ the weight assigned to $X_j$ at point $\inp$. Let $v: 2^d \mapsto \mathbb{R}$ be a value function such that $v(S)$ is the payoff associated with feature subset $S \subseteq [d]$ and $v(\{\emptyset\}) = 0$. Define the complement $R = [d] \backslash S$ such that we may rewrite any $\inp$ as a pair of subvectors, $(\bm{x}_i^S, \bm{x}_i^R)$. Payoffs are given by:
\begin{equation}
\label{eq:val}
    v(S) = \mathop{\mathbb{E}}[f(\bm{x}_i^S, \bm{X}^R)],
\end{equation}
although this introduces some ambiguity regarding the reference distribution for $\bm{X}^R$ (more on this below). The Shapley value $\phi_j$ is then $j$'s average marginal contribution to all subsets that exclude it:
\begin{equation} \label{eq:shapley}
    \phi_j = \sum_{S \subseteq [d] \backslash \{j\}} \frac{|S|!(d - |S| - 1)!}{d!} v(S \cup \{j\}) - v(S).
\end{equation}
It can be shown that this is the unique solution to the attribution problem that satisfies certain desirable properties, including efficiency, linearity, sensitivity, and symmetry. 

Reformulating this in our framework, we find that the value function $v$ is a sufficiency measure. To see this, let each $\bm{z} \sim \mathcal{D}$ be a sample in which a random subset of variables $S$ are held at their original values, while remaining features $R$ are drawn from a fixed distribution $\mathcal{D}(\cdot|S)$.\footnote{The diversity of Shapley value algorithms is largely due to variation in how this distribution is defined. Popular choices include the marginal $P(\bm{X}^R)$ \citep{lundberg_lee_2017}; conditional $P(\bm{X}^R|\bm{x}^S)$ \citep{Aas2019}; and interventional $P(\bm{X}^R|do(\bm{x}^S))$ \citep{heskes2020} distributions.} 

\begin{prop} \label{prop:shap1}
    Let $c_S(\bm{z}) = 1$ iff $\bm{x} \subseteq \bm{z}$ was constructed by holding $\bm{x}^S$ fixed and sampling $\bm{X}^R$ according to $\mathcal{D}(\cdot|S)$. Then $v(S) = PS(c_S, y)$.
\end{prop}
Thus, the Shapley value $\phi_j$ measures $X_j$'s average marginal increase to the sufficiency of a random feature subset. The advantage of our method is that, by focusing on particular subsets instead of weighting them all equally, we disregard irrelevant permutations and home in on just those that meet a $\tau$-minimality criterion. \citet{kumar2020} observe that, ``since there is no standard procedure for converting Shapley values into a statement about a model’s behavior, developers rely on their own mental model of what the values represent'' (p. 8). By contrast, necessary and sufficient factors are more transparent and informative, offering a direct path to what Shapley values indirectly summarize. 


\paragraph{Rule lists.}\label{sec:rule_list} 
Rule lists are sequences of if-then statements that describe a hyperrectangle in feature space, creating partitions that can be visualized as decision or regression trees. Rule lists have long been popular in XAI. While early work in this area tended to focus on global methods \citep{Friedman2008, Letham2015}, more recent efforts have prioritized local explanation tasks \citep{lakkaraju2019, sokol2020limetree}. 

We focus in particular on the Anchors algorithm \citep{Ribeiro2018}, which learns a set of Boolean conditions $A$ (the eponymous ``anchors'') such that $A(\inp) = 1$ and 
\begin{equation} \label{eq:anchors}
    P_{\mathcal{D}_{(\bm{x}|A)}}(f(\inp) = f(\bm{x})) \geq \tau. 
\end{equation}
The lhs of Eq.~\ref{eq:anchors} is termed the \emph{precision}, prec($A$), and probability is taken over a synthetic distribution in which the conditions in $A$ hold while other features are perturbed. Once $\tau$ is fixed, the goal is to maximize \emph{coverage}, formally defined as $\mathds{E}[A(\cfinp) = 1]$, i.e. the proportion of datapoints to which the anchor applies. 

The formal similarities between Eq.~\ref{eq:anchors} and Def.~\ref{def:probsuff} are immediately apparent, and the authors themselves acknowledge that Anchors are intended to provide ``sufficient conditions'' for model predictions. 

\begin{prop} \label{prop:anchors}
    Let $c_A(\bm{z}) = 1$ iff $A(\bm{x}) = 1$. Then $\text{prec}(A) = PS(c_A, y)$.
\end{prop}

While Anchors outputs just a single explanation, our method generates a ranked list of candidates, thereby offering a more comprehensive view of model behavior. Moreover, our necessity measure adds a mode of explanatory information entirely lacking in Anchors. 

\paragraph{Counterfactuals.}\label{sec:counterfactuals} 
Counterfactual explanations identify one or several nearest neighbors with different outcomes, e.g. all datapoints $\cfinp$ within an $\epsilon$-ball of $\inp$ such that labels $f(\cfinp)$ and $f(\inp)$ differ (for classification) or $f(\cfinp) > f(\inp) + \delta$ (for regression).\footnote{Confusingly, the term ``counterfactual'' in XAI refers to any point with an alternative outcome, which is distinct from the causal sense of the term (see Sect.~\ref{sec:nec_suf}). We use the word in both senses here, but strive to make our intended meaning explicit in each case.}
The optimization problem is: 
\begin{equation} \label{eq:cf}
    \bm{x}^* = \argmin_{\cfinp \in \text{CF}(\inp)} ~cost(\inp, \cfinp),
\end{equation}
where $\text{CF}(\inp)$ denotes a counterfactual space such that $f(\inp) \neq f(\cfinp)$ and $cost$ is a user-supplied cost function, typically equated with some distance measure. \citep{Wachter2018} recommend using generative adversarial networks to solve Eq.~\ref{eq:cf}, while others have proposed alternatives designed to ensure that counterfactuals are coherent and actionable \citep{ustun_actionable, karimi2020survey, wexler_whatif}.
As with Shapley values, the variation in these proposals is reducible to the choice of context $\mathcal{D}$.

For counterfactuals, we rewrite the objective as a search for minimal perturbations sufficient to flip an outcome.

\begin{table*}[!t]
\caption{Overview of experimental settings by basis configuration.}
\vspace{-2ex}
\centering
\resizebox{\textwidth}{!}{%
\begin{tabular}{@{}|llllll|@{}}
\toprule
\rowcolor[HTML]{C0C0C0} 
\cellcolor[HTML]{FE996B}Experiment &
  Datasets &
  $f$ &
  $\mathcal{D}$ &
  $\mathcal{C}$ &
  $\preceq$ \\ \midrule
\multicolumn{1}{|l|}{Attribution comparison} &
  \multicolumn{1}{l|}{\texttt{German}, \texttt{SpamAssassins}} &
  \multicolumn{1}{l|}{\texttt{Extra-Trees}} &
  \multicolumn{1}{l|}{R2I, I2R} &
  \multicolumn{1}{l|}{Intervention targets} &
  - \\
\multicolumn{1}{|l|}{Anchors comparison: Brittle predictions} &
  \multicolumn{1}{l|}{\texttt{IMDB}} &
  \multicolumn{1}{l|}{\texttt{LSTM}} &
  \multicolumn{1}{l|}{R2I, I2R} &
  \multicolumn{1}{l|}{Intervention targets} &
  $\preceq_{subset}$ \\
\multicolumn{1}{|l|}{Anchors comparison: PS and Prec} &
  \multicolumn{1}{l|}{\texttt{German}} &
  \multicolumn{1}{l|}{\texttt{Extra-Trees}} &
  \multicolumn{1}{l|}{R2I} &
  \multicolumn{1}{l|}{Intervention targets} &
  $\preceq_{subset}$ \\
\multicolumn{1}{|l|}{Counterfactuals: Adverserial} &
  \multicolumn{1}{l|}{\texttt{SpamAssassins}} &
  \multicolumn{1}{l|}{\texttt{MLP}} &
  \multicolumn{1}{l|}{R2I} &
  \multicolumn{1}{l|}{Intervention targets} &
  $\preceq_{subset}$ \\
\multicolumn{1}{|l|}{Counterfactuals: Recourse, DiCE comparison} &
  \multicolumn{1}{l|}{\texttt{Adult}} &
  \multicolumn{1}{l|}{\texttt{MLP}} &
  \multicolumn{1}{l|}{I2R} &
  \multicolumn{1}{l|}{Full interventions} &
  $\preceq_{cost}$ \\
\multicolumn{1}{|l|}{Counterfactuals: Recourse, causal vs. non-causal} &
  \multicolumn{1}{l|}{\texttt{German}} &
  \multicolumn{1}{l|}{\texttt{Extra-Trees}} &
  \multicolumn{1}{l|}{I2R$_{causal}$} &
  \multicolumn{1}{l|}{Full interventions} &
  $\preceq_{cost}$ \\ \bottomrule
\end{tabular}%
}
\vspace{-2ex}
\label{tab:expriments}
\end{table*}

\begin{prop} \label{prop:cf}
    Let $cost$ be a function representing $\preceq$, and let $c$ be some factor spanning reference values. Then the counterfactual recourse objective is:
    \begin{equation}
        c^* = \argmin_{c \in \mathcal{C}} ~cost(c) ~~\textrm{s.t.} ~PS(c, 1 - y) \geq \tau,
    \end{equation}
    where $\tau$ denotes a decision threshold. Counterfactual outputs will then be any $\bm{z} \sim \mathcal{D}$ such that $c^*(\bm{z}) = 1$. 
\end{prop}


\paragraph{Probabilities of causation.}\label{sec:prob_causation}
Our framework can describe \citet{Pearl2000}'s aforementioned probabilities of causation, however in this case $\mathcal{D}$ must be constructed with care. 

\begin{prop}\label{prop:pearl}
    Consider the bivariate Boolean setting, as in Sect.~\ref{sec:nec_suf}. We have two counterfactual distributions: an input space $\mathcal{I}$, in which we observe $x, y$ but intervene to set $X = x'$; and a reference space $\mathcal{R}$, in which we observe $x', y'$ but intervene to set $X = x$. Let $\mathcal{D}$ denote a uniform mixture over both spaces, and let auxiliary variable $W$ tag each sample with a label indicating whether it comes from the original ($W = 1$) or contrastive ($W = 0$) counterfactual space. Define $c(\bm{z}) = w$. Then we have $\texttt{suf}(x, y) = PS(c, y)$ and $\texttt{nec}(x, y) = PS(1 - c, y')$.
\end{prop}
In other words, we regard Pearl's notion of necessity as \emph{sufficiency of the negated factor for the alternative outcome}. By contrast, \citet{Pearl2000} has no analogue for our probability of necessity. This is true of any measure that defines sufficiency and necessity via inverse, rather than converse probabilities. While conditioning on the same variable(s) for both measures may have some intuitive appeal, it comes at a cost to expressive power. Whereas our framework can recover all four explanatory measures, corresponding to the classical definitions and their contrapositive forms, definitions that merely negate instead of transpose the antecedent and consequent are limited to just two. 


\begin{remark}
    We have assumed that factors and outcomes are Boolean throughout. Our results can be extended to continuous versions of either or both variables, so long as $c(\bm{Z})~ \indep ~Y~|~\bm{Z}$. This conditional independence holds whenever $\bm{W}~ \indep ~Y~|~\bm{X}$, which is true by construction since $f(\bm{z}) := f(\bm{x})$. However, we defend the Boolean assumption on the grounds that it is well motivated by contrastivist epistemologies \citep{Kahneman1986, lipton_1990, Blaauw2013} and not especially restrictive, given that partitions of arbitrary complexity may be defined over $\bm{Z}$ and $Y$. 
\end{remark}

\section{Experiments}\label{sec:experiments}


\begin{figure}[!t]
    \centering
    \includegraphics[scale=0.35]{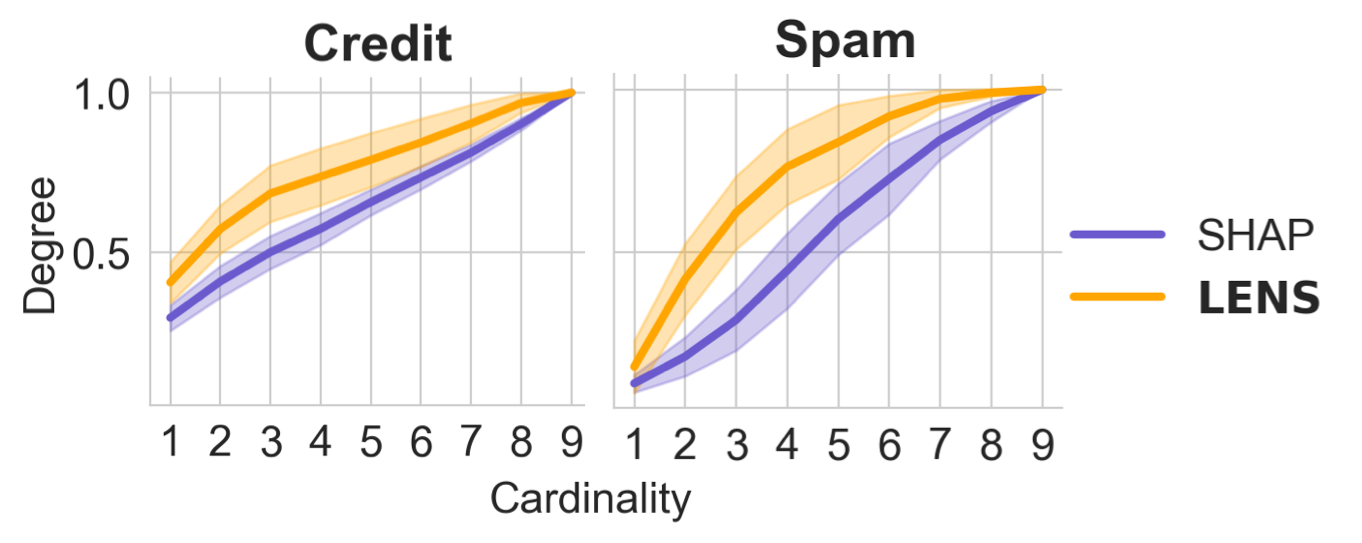}
    \vspace{-2ex}
    \caption{Comparison of top $k$ features ranked by SHAP against the best performing LENS subset of size $k$ in terms of $PS(c,y)$. \texttt{German} results are over 50 inputs; \texttt{SpamAssassins} results are over 25 inputs.}
    \vspace{-2ex}
    \label{fig:shap_compare}
\end{figure}

In this section, we demonstrate the use of LENS on a variety of tasks and compare results with popular XAI tools, using the basis configurations detailed in Table \ref{tab:expriments}. A comprehensive discussion of experimental design, including datasets and pre-processing pipelines, is left to Appendix \ref{sec:app_exp}. Code for reproducing all results is available at \url{https://github.com/limorigu/LENS}.


\vspace{-2.5mm}
\paragraph{Contexts.}
We consider a range of contexts $\mathcal{D}$ in our experiments. For the input-to-reference (I2R) setting, we replace input values with reference values for feature subsets $S$; for the reference-to-input (R2I) setting, we replace reference values with input values. 
We use R2I for examining sufficiency/necessity of the original model prediction, and I2R for examining sufficiency/necessity of a contrastive model prediction.
We sample from the empirical data in all experiments, except in Sect.~\ref{sec:exp_causal}, where we assume access to a structural causal model (SCM). 


\vspace{-2.5mm}
\paragraph{Partial Orderings.}
We consider two types of partial orderings in our experiments. The first, $\preceq_{subset}$, evaluates subset relationships. For instance, if $c(\bm{z}) = \mathds{1}[\bm{x}[\mathsf{gender} = \text{``female''}]]$ and $c'(\bm{z}) = \mathds{1}[\bm{x}[\mathsf{gender} = \text{``female''} \land \mathsf{age \geq 40}]]$, then we say that $c \preceq_{subset} c'$. 
The second, $c \preceq_{cost} c' := c \preceq_{subset} c' ~\wedge~ cost(c) \leq cost(c')$, adds the additional constraint that $c$ has cost no greater than $c'$.
The cost function could be arbitrary.
Here, we consider distance measures over either the entire state space or just the intervention targets corresponding to $c$.

\subsection{Feature Attributions}\label{sec:exp_attributions}
\vspace{-2.5mm}
Feature attributions are often used to identify the top-$k$ most important features for a given model outcome~\citep{barocas2020}. However, we argue that these feature sets may not be explanatory with respect to a given prediction.
To show this, we compute R2I and I2R sufficiency -- i.e., $PS(c, y)$ and $PS(1 - c, 1 - y)$, respectively -- for the top-$k$ most influential features ($k \in [1,9]$) as identified by SHAP \citep{lundberg_lee_2017} and LENS. 
Fig. \ref{fig:shap_compare} shows results from the R2I setting for \texttt{German} credit \citep{UCI_Dua:2019} and \texttt{SpamAssassin} datasets \citep{SpamDataset}. 
Our method attains higher $PS$ for all cardinalities. We repeat the experiment over 50 inputs, plotting means and 95\% confidence intervals for all $k$. 
Results indicate that our ranking procedure delivers more informative explanations than SHAP at any fixed degree of sparsity. 
Results from the I2R setting are in Appendix \ref{sec:app_exp}.

\begin{table*}[t!]
\caption{Example prediction given by an LSTM model trained on the \texttt{IMDB} dataset. We compare $\tau$-minimal factors identified by LENS (as individual words), based on $PS(c, y)$ and $PS(1-c, 1-y)$, and compare to output by Anchors. 
}
\vspace{-2ex}
\centering
\resizebox{\textwidth}{!}{%
\begin{tabular}{@{}|ll|ll|ll|@{}}
\toprule
\rowcolor[HTML]{C0C0C0} 
\multicolumn{2}{|l|}{\cellcolor[HTML]{FE996B}Inputs} &
  \multicolumn{2}{l|}{\cellcolor[HTML]{C0C0C0}Anchors} &
  \multicolumn{2}{l|}{\cellcolor[HTML]{C0C0C0}\textbf{LENS}} \\ \midrule
\multicolumn{1}{|l|}{\cellcolor[HTML]{EFEFEF}{Text}} &
  {\cellcolor[HTML]{EFEFEF}Original model prediction} &
  \multicolumn{1}{l|}{\cellcolor[HTML]{EFEFEF}{Suggested anchors}} &
  {\cellcolor[HTML]{EFEFEF}Precision} &
  \multicolumn{1}{l|}{\cellcolor[HTML]{EFEFEF}{Sufficient R2I factors}} &
  {\cellcolor[HTML]{EFEFEF}Sufficient I2R factors} \\ \midrule
\rowcolor[HTML]{FFFFFF} 
'read book forget movie' &
  {\color[HTML]{CB0000} wrongly predicted positive} &
  [read, movie] &
  0.94 &
  [read, forget, movie] &
  read, forget, movie \\
\rowcolor[HTML]{FFFFFF} 
'you better choose paul verhoeven even watched' &
  {\color[HTML]{036400} correctly predicted negative} &
  [choose, better, even, you, paul, verhoeven] &
  0.95 &
  choose, even &
  better, choose, paul, even \\
  \bottomrule
\end{tabular}%
}
\vspace{-1ex}
\label{tab:sentiment_weakness}
\end{table*}

\begin{table*}[t!]
    \caption{(Top) A selection of emails from \texttt{SpamAssassins}, correctly identified as spam by an MLP. The goal is to find minimal perturbations that result in non-spam predictions. (Bottom) Minimal subsets of feature-value assignments that achieve non-spam predictions with respect to the emails above.}
    \vspace{-2ex}
\centering
\resizebox{\textwidth}{!}{%
\begin{tabular}{@{}|l|l|l|l|l|@{}}
\toprule
\rowcolor[HTML]{C0C0C0} 
From                  & To                                            & Subject                                       & First Sentence & Last Sentence                   \\ \midrule
resumevalet info resumevalet com &
  yyyy cv spamassassin taint org &
  adv put resume back work &
  dear candidate &
  professionals online network inc \\
jacqui devito goodroughy ananzi co za &
  picone linux midrange com &
  enlargement breakthrough zibdrzpay &
  recent survey conducted &
  increase size enter detailsto come open \\
rose xu email com     & yyyyac idt net                                & adv harvest lots target email address quickly & want           & advertisement persons 18yrs old \\ \bottomrule
\end{tabular}%
\vspace{-3ex}}\label{tab:spam_game_refs}

\vspace{-4ex}
\end{table*}

\begin{table}[t!]
    \resizebox{0.5\textwidth}{!}{%
\begin{tabular}{|l|ll|}
\toprule
\cellcolor[HTML]{FE996B}Gaming options &
  \multicolumn{2}{l|}{\cellcolor[HTML]{C0C0C0} Feature subsets for value changes} \\ \midrule
                    & \multicolumn{1}{l|}{\cellcolor[HTML]{EFEFEF}From} & \cellcolor[HTML]{EFEFEF}To             \\ \cmidrule(l){2-3} 
\multirow{-2}{*}{1} &
  crispin cown crispin wirex com &
  example com mailing... list secprog securityfocus... moderator\\ \cmidrule(r){1-1}
                    & \multicolumn{1}{l|}{\cellcolor[HTML]{EFEFEF}From} & \cellcolor[HTML]{EFEFEF}First Sentence \\ \cmidrule(l){2-3} 
\multirow{-2}{*}{2} & crispin cowan crispin wirex com                   & scott mackenzie wrote                  \\ \cmidrule(r){1-1}
                    & \multicolumn{1}{l|}{\cellcolor[HTML]{EFEFEF}From} & \cellcolor[HTML]{EFEFEF}First Sentence \\ \cmidrule(l){2-3} 
\multirow{-2}{*}{3} & tim one comcast net tim peters                    & tim                                    \\ \bottomrule
\end{tabular}

    }
    \vspace{-2.5ex}
\end{table}

\subsection{Rule Lists}

\paragraph{Sentiment sensitivity analysis.}
Next, we use LENS to study model weaknesses by considering minimal factors with high R2I and I2R sufficiency in text models. Our goal is to answer questions of the form, ``What are words with/without which our model would output the original/opposite prediction for an input sentence?''
For this experiment, we train an LSTM network on the \texttt{IMDB} dataset for sentiment analysis \citep{maas-EtAl2011}. 
If the model mislabels a sample, we investigate further; if it does not, we inspect the most explanatory factors to learn more about model behavior. For the purpose of this example, we only inspect sentences of length 10 or shorter.
We provide two examples below and
compare with Anchors (see Table \ref{tab:sentiment_weakness}).

\begin{table*}[t!]
\caption{Recourse example comparing causal and non-causal (i.e., feature independent) $\mathcal{D}$. We sample a single input example with a negative prediction, and 100 references with the opposite outcome. 
For I2R$_{causal}$ we propagate the effects of interventions through a user-provided SCM. 
}
\vspace{-2ex}
\centering
\resizebox{\textwidth}{!}{%
\begin{tabular}{@{}|lllllllllllll|@{}}
\toprule
\rowcolor[HTML]{FE996B} 
\multicolumn{9}{|l|}{\cellcolor[HTML]{FE996B}input} &
  \multicolumn{2}{l}{\cellcolor[HTML]{C0C0C0}I2R} &
  \multicolumn{2}{l|}{\cellcolor[HTML]{C0C0C0}I2R$_{causal}$} \\ \midrule
\rowcolor[HTML]{EFEFEF}
\multicolumn{1}{|l|}{\cellcolor[HTML]{EFEFEF}Age} &
  \multicolumn{1}{l|}{\cellcolor[HTML]{EFEFEF}Sex} &
  \multicolumn{1}{l|}{\cellcolor[HTML]{EFEFEF}Job} &
  \multicolumn{1}{l|}{\cellcolor[HTML]{EFEFEF}Housing} &
  \multicolumn{1}{l|}{\cellcolor[HTML]{EFEFEF}Savings} &
  \multicolumn{1}{l|}{\cellcolor[HTML]{EFEFEF}Checking} &
  \multicolumn{1}{l|}{\cellcolor[HTML]{EFEFEF}Credit} &
  \multicolumn{1}{l|}{\cellcolor[HTML]{EFEFEF}Duration} &
  \multicolumn{1}{l|}{\cellcolor[HTML]{EFEFEF}Purpose} &
  \multicolumn{1}{l|}{\cellcolor[HTML]{EFEFEF}$\tau$-minimal factors ($\tau=0$)} &
  \multicolumn{1}{l|}{\cellcolor[HTML]{EFEFEF}Cost} &
  \multicolumn{1}{l|}{\cellcolor[HTML]{EFEFEF}$\tau$-minimal factors ($\tau=0$)} &
  Cost \\ \midrule
 &
   &
   &
   &
   &
   &
   &
   &
  \multicolumn{1}{l|}{} &
  Job: Highly skilled &
  \multicolumn{1}{l|}{1} &
  Age: 24 &
  0.07 \\
 &
   &
   &
   &
   &
   &
   &
   &
  \multicolumn{1}{l|}{} &
  Checking: NA &
  \multicolumn{1}{l|}{1} &
  Sex: Female &
  1 \\
 &
   &
   &
   &
   &
   &
   &
   &
  \multicolumn{1}{l|}{} &
  Duration: 30 &
  \multicolumn{1}{l|}{1.25} &
  Job: Highly skilled &
  1 \\
 &
   &
   &
   &
   &
   &
   &
   &
  \multicolumn{1}{l|}{} &
  Age: 65, Housing: Own &
  \multicolumn{1}{l|}{4.23} &
  Housing: Rent &
  1 \\
\multirow{-5}{*}{23} &
  \multirow{-5}{*}{Male} &
  \multirow{-5}{*}{Skilled} &
  \multirow{-5}{*}{Free} &
  \multirow{-5}{*}{Little} &
  \multirow{-5}{*}{Little} &
  \multirow{-5}{*}{1845} &
  \multirow{-5}{*}{45} &
  \multicolumn{1}{l|}{\multirow{-5}{*}{Radio/TV}} &
  Age: 34, Savings: N/A &
  \multicolumn{1}{l|}{1.84} &
  Savings: N/A &
  1 \\ \bottomrule
\end{tabular}%
}
\vspace{-1ex}
\label{tab:causal_noncausal_recourse}
\end{table*}

Consider our first example: \textsc{read book forget movie} is a sentence we would expect to receive a negative prediction, but our model classifies it as positive. Since we are investigating a positive prediction, our reference space is conditioned on a negative label. For this model, the classic \textsc{UNK} token receives a positive prediction. Thus we opt for an alternative, \textsc{PLATE}. Performing interventions on all possible combinations of words with our token, we find the conjunction of \textsc{read, forget}, and \textsc{movie} is a sufficient factor for a positive prediction (R2I). We also find that changing any of \textsc{read}, \textsc{forget}, or \textsc{movie} to PLATE would result in a negative prediction (I2R). Anchors, on the other hand, perturbs the data stochastically (see Appendix \ref{sec:app_exp}), suggesting the conjunction \textsc{read AND book}. Next, we investigate the sentence: \textsc{you better choose paul verhoeven even watched}. Since the label here is negative, we use the \textsc{UNK} token. We find that this prediction is brittle -- a change of almost any word would be sufficient to flip the outcome. Anchors, on the other hand, reports a conjunction including most words in the sentence. Taking the R2I view, we still find a more concise explanation: \textsc{choose} or \textsc{even} would be enough to attain a negative prediction. These brief examples illustrate how LENS may be used to find brittle predictions across samples, search for similarities between errors, or test for model reliance on sensitive attributes (e.g., gender pronouns).

\vspace{-2mm}
\paragraph{Anchors comparison.} Anchors also includes a tabular variant, against which we compare LENS's performance in terms of R2I sufficiency. We present the results of this comparison in Fig.~\ref{fig:anchors_mean_suff_prec}, and include additional comparisons in Appendix \ref{sec:app_exp}. We sample 100 inputs from the \texttt{German} dataset, and query both methods with $\tau=0.9$ using the classifier from Sect.~\ref{sec:exp_attributions}. Anchors satisfies a PAC bound controlled by parameter $\delta$. At the default value $\delta=0.1$, Anchors fails 
to meet the $\tau$ threshold on 14\% of samples; LENS meets it on 100\% of samples. This result accords with Thm.~\ref{thm:complete}, and vividly demonstrates the benefits of our optimality guarantee. Note that we also go beyond Anchors in providing multiple explanations instead of just a single output, as well as a cumulative probability measure with no analogue in their algorithm.


\begin{figure}[!t]
    \vspace{1ex}
    \centering
    \includegraphics[scale=0.35]{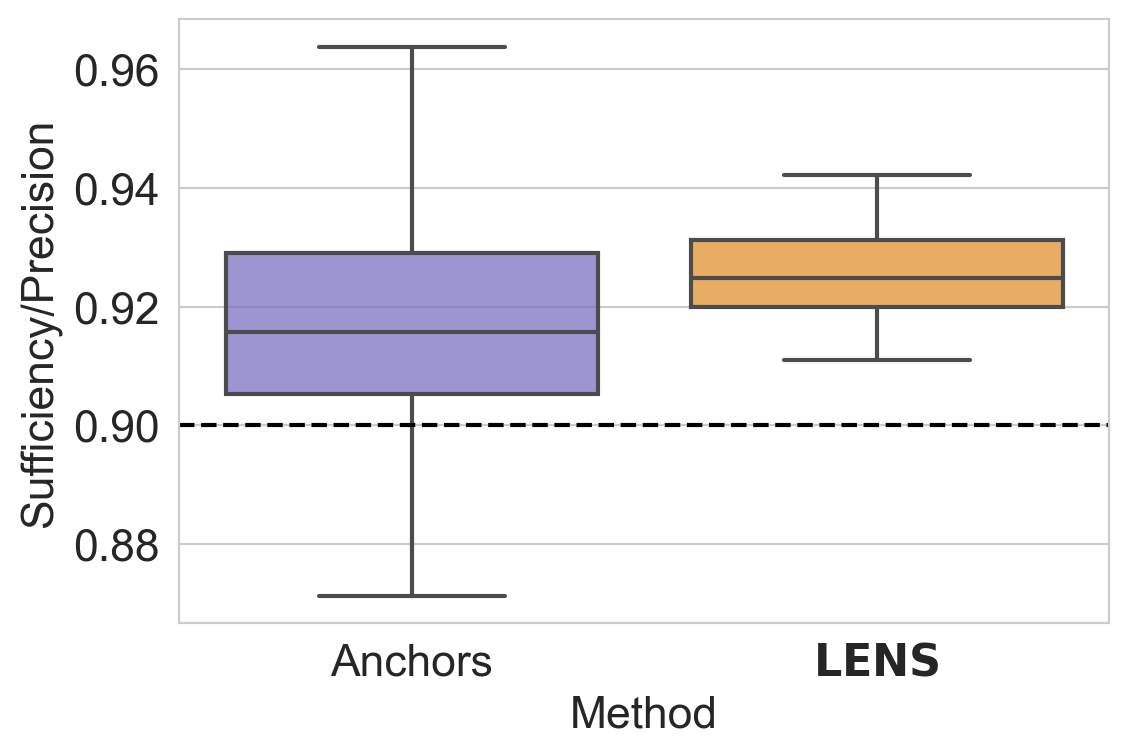}
    \caption{We compare $PS(c, y)$ against precision scores attained by the output of LENS and Anchors for examples from \texttt{German}. We repeat the experiment for 100 inputs, and each time consider the single example generated by Anchors against the mean $PS(c, y)$ among LENS's candidates. Dotted line indicates $\tau=0.9$.}
    \vspace{-3.5ex}
    \label{fig:anchors_mean_suff_prec}
\end{figure}
\subsection{Counterfactuals}\label{subset:counterfactuals}
\paragraph{Adversarial examples: spam emails.}
 
R2I sufficiency answers questions of the form, ``What would be sufficient for the model to predict $y$?''. This is particularly valuable in cases with unfavorable outcomes $y'$. Inspired by adversarial interpretability approaches \citep{ribeiro2018semantically, lakk_fool_lime}, we train an MLP classifier on the \texttt{SpamAssassins} dataset and search for minimal factors sufficient to relabel a sample of spam emails as non-spam.
Our examples follow some patterns common to spam emails: received from unusual email addresses, includes suspicious keywords such as \textsc{enlargement} or \textsc{advertisement} in the subject line, etc. We identify minimal changes that will flip labels to non-spam with high probability. Options include altering the incoming email address to more common domains, and changing the subject or first sentences (see Table \ref{tab:spam_game_refs}). These results can improve understanding of both a model's behavior and a dataset's properties.

\vspace{-2.5mm}
\paragraph{Diverse counterfactuals.}\label{sec:dice}
Our explanatory measures can also be used to secure algorithmic recourse. 
For this experiment, we benchmark against DiCE \citep{mothilal2020_dice}, which aims to provide diverse recourse options for any underlying prediction model. We illustrate the differences between our respective approaches on the \texttt{Adult} dataset \citep{AdultInc96}, using an MLP and following the procedure from the original DiCE paper. 

According to DiCE, a diverse set of counterfactuals is one that differs in \textit{values} assigned to features, and can thus produce a counterfactual set that includes different interventions on the same variables (e.g., CF1: $\mathsf{age} = 91, \mathsf{occupation}$ = ``retired''; CF2: $\mathsf{age} = 44, \mathsf{occupation}$ = ``teacher'').
Instead, we look at diversity of counterfactuals in terms of intervention \emph{targets}, i.e. features changed (in this case, from input to reference values) and their effects.
We present minimal cost interventions that would lead to recourse for each feature set but we summarize the set of paths to recourse via subsets of features changed. 
Thus, DiCE provides answers of the form ``Because you are not 91 and retired'' or ``Because you are not 44 and a teacher''; we answer ``Because of your age and occupation'', and present the lowest cost intervention on these features sufficient to flip the prediction. 

\begin{figure}
    \vspace{2ex}
    \centering
    \includegraphics[scale=0.25]{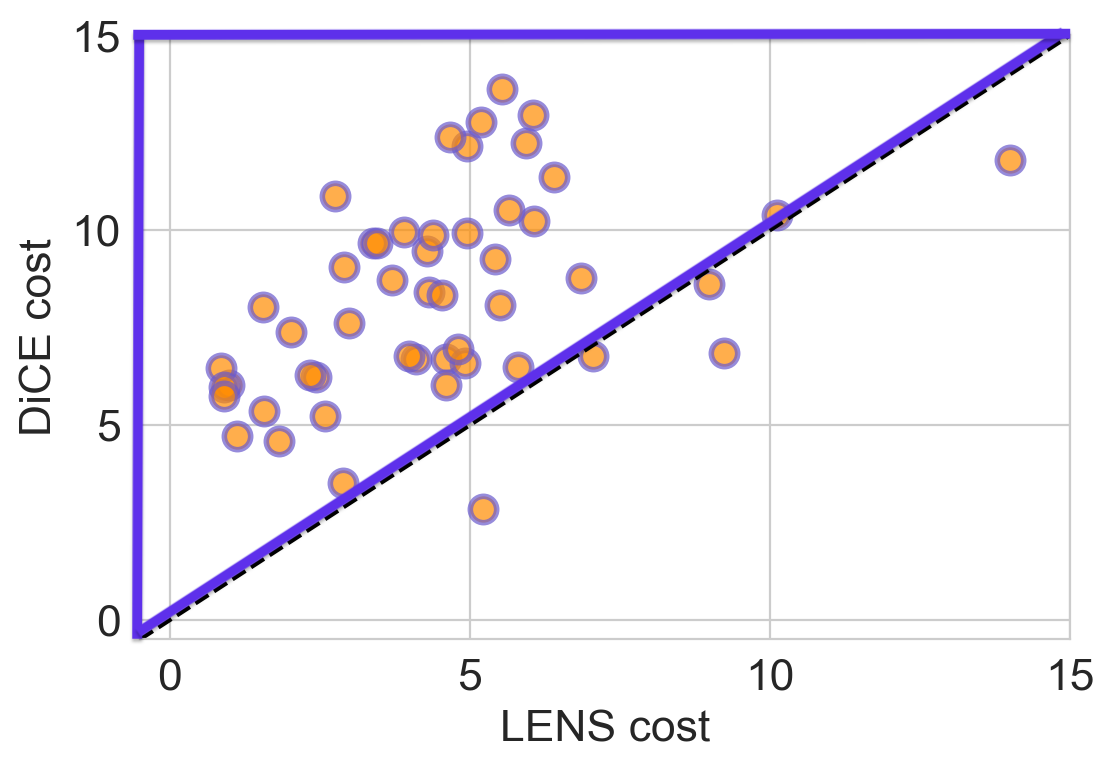}
    \vspace{-3ex}
    \caption{A comparison of mean cost of outputs by LENS and DiCE for 50 inputs sampled from the \texttt{Adult} dataset.}
    \vspace{-3.5ex}
    \label{fig:recourse_DiCE_cost}
\end{figure}

With this intuition in mind, we compare outputs given by DiCE and LENS for various inputs. For simplicity, we let all features vary independently. 
We consider two metrics for comparison: (a) the mean cost of proposed factors, 
and (b) the number of minimally valid candidates proposed, where a factor $c$ from a method $M$ is \emph{minimally valid} iff for all $c'$ proposed by $M'$, $\neg{(c' \prec_{cost} c)}$ (i.e., $M'$ does not report a factor  preferable to $c$).
We report results based on 50 randomly sampled inputs from the \texttt{Adult} dataset, where references are fixed by conditioning on the opposite prediction. The cost comparison results are shown in Fig. \ref{fig:recourse_DiCE_cost}, where we find that LENS identifies lower cost factors for the vast majority of inputs.
Furthermore, DiCE finds no minimally valid candidates that LENS did not already account for.
Thus LENS emphasizes \textit{minimality} and \textit{diversity} of intervention targets, while still identifying low cost intervention values. 

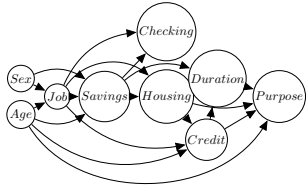
\begin{figure}
    \vspace{-1ex}
    \centering
    \resizebox{0.23\textwidth}{!}{\begin{tikzpicture}
  \node[obs] (Age) at (-4,0) {$Age$};
  \node[obs] (Sex) at (-4,1) {$Sex$};
  \node[obs] (Job) at (-3, 0.5) {$Job$};
  \node[obs] (Save) at (-1.7, 0.5) {$Savings$};
  \node[obs] (Housing) at (0, 0.5) {$Housing$};
  \node[obs] (Checking) at (0, 2.3) {$Checking$};
  \node[obs] (Credit) at (1.1, -0.7) {$Credit$};
  \node[obs] (Duration) at (1.4, 1) {$Duration$};
  \node[obs] (Purpose) at (3.1, 0.5) {$Purpose$};
  \edge{Age}{Job};
  \edge[bend right=30]{Age}{Save};
  \edge[bend right=35]{Age}{Credit};
  \edge[bend right=60]{Age}{Purpose};
  \edge{Sex}{Job};
  \edge[bend left]{Sex}{Save};
  \edge{Job}{Save};
  \edge[bend left]{Job}{Checking};
  \edge[bend left=50]{Job}{Housing};
  \edge[bend right]{Job}{Credit};
  \edge{Save}{Checking};
  \edge[bend left]{Save}{Duration};
  \edge{Save}{Housing};
  \edge{Housing}{Credit};
  \edge[bend right=15]{Housing}{Purpose};
  \edge{Credit}{Duration};
  \edge{Credit}{Purpose};
  \edge{Duration}{Purpose};
  \end{tikzpicture}}
    \caption{Example DAG for \texttt{German} dataset.}
    \label{fig:recourse_DAG}
    \vspace{-1ex}
\end{figure}

\vspace{-2.5mm}
\paragraph{Causal vs. non-causal recourse.}\label{sec:exp_causal}
When a user relies on XAI methods to plan interventions on real-world systems, causal relationships between predictors cannot be ignored. In the following example, we consider the DAG in Fig. \ref{fig:recourse_DAG}, intended to represent dependencies in the \texttt{German} credit dataset. For illustrative purposes, we assume access to the structural equations of this data generating process. (There are various ways to extend our approach using only partial causal knowledge as input \citep{karimi2020imperf, heskes2020}.) We construct $D$ by sampling from the SCM under a series of different possible interventions. Table \ref{tab:causal_noncausal_recourse} describes an example of how using our framework with augmented causal knowledge can lead to different recourse options. Computing explanations under the assumption of feature independence results in factors that span a large part of the DAG depicted in Fig. \ref{fig:recourse_DAG}. However, encoding structural relationships in $D$, we find that LENS assigns high explanatory value to nodes that appear early in the topological ordering. 
This is because intervening on a single root factor may result in various downstream changes once effects are fully propagated. 

\section{Discussion}\label{sec:discussion}
Our results, both theoretical and empirical, rely on access to the relevant context $\mathcal{D}$ and the complete enumeration of all feature subsets. Neither may be feasible in practice. When elements of $\bm{Z}$ are estimated, as is the case with the generative methods sometimes used in XAI, modeling errors could lead to suboptimal explanations. For high-dimensional settings such as image classification, LENS cannot be na{\"i}vely applied without substantial data pre-processing. The first issue is extremely general. No method is immune to model misspecification, and attempts to recreate a data generating process must always be handled with care. Empirical sampling, which we rely on above, is a reasonable choice when data are fairly abundant and representative. However, generative models may be necessary to correct for known biases or sample from low-density regions of the feature space. This comes with a host of challenges that no XAI algorithm alone can easily resolve. 
The second issue -- that a complete enumeration of all variable subsets is often impractical -- we consider to be a feature, not a bug. Complex explanations that cite many contributing factors pose \emph{cognitive} as well as computational challenges. In an influential review of XAI, \cite{Miller_2019} finds near unanimous consensus among philosophers and social scientists that, ``all things being equal, simpler explanations -- those that cite fewer causes... are better explanations'' (p. 25). Even if we could list all $\tau$-minimal factors for some very large value of $d$, it is not clear that such explanations would be helpful to humans, who famously struggle to hold more than seven objects in short-term memory at any given time \citep{Miller1955}. That is why many popular XAI tools include some sparsity constraint to encourage simpler outputs.

Rather than throw out some or most of our low-level features, we prefer to consider a higher level of abstraction, where explanations are more meaningful to end users. For instance, in our \texttt{SpamAssassins} experiments, we started with a pure text example, which can be represented via high-dimensional vectors (e.g., word embeddings). However, we represent the data with just a few intelligible components: \texttt{From} and \texttt{To} email addresses, \texttt{Subject}, etc. In other words, we create a more abstract object and consider each segment as a potential intervention target, i.e. a candidate factor. This effectively compresses a high-dimensional dataset into a 10-dimensional abstraction. Similar strategies could be used in many cases, either through domain knowledge or data-driven clustering and dimensionality reduction techniques \citep{Chalupka2017, Beckers2019, locatello19a}. In general, if data cannot be represented by a reasonably low-dimensional, intelligible abstraction, then post-hoc XAI methods are unlikely to be of much help.
\section{Conclusion}\label{sec:conclusion}
We have presented a unified framework for XAI that foregrounds necessity and sufficiency, which we argue are the fundamental building blocks of all successful explanations. We defined simple measures of both, and showed how they undergird various XAI methods. Our formulation, which relies on converse rather than inverse probabilities, is uniquely flexible and expressive. It covers all four basic explanatory measures -- i.e., the classical definitions and their contrapositive transformations -- and unambiguously accommodates logical, probabilistic, and/or causal interpretations, depending on how one constructs the basis tuple $\mathcal{B}$. We illustrated illuminating connections between our measures and existing proposals in XAI, as well as \citet{Pearl2000}'s probabilities of causation. We introduced a sound and complete algorithm for identifying minimally sufficient factors, and demonstrated our method on a range of tasks and datasets. Our approach prioritizes completeness over efficiency, suitable for settings of moderate dimensionality. Future research will explore more scalable approximations, model-specific variants optimized for, e.g., convolutional neural networks, and developing a graphical user interface. 


\begin{acknowledgements} 
    DSW was supported by ONR grant N62909-19-1-2096.
\end{acknowledgements}








\bibliography{references}
\linepenalty=3000
\appendix 

\renewcommand{\thesection}{\Alph{section}}

\section{Proofs}\label{sec:app_proofs}

\subsection{Theorems}

\subsubsection{Proof of Theorem \ref{thm:complete}}

\begin{theorem*}
    With oracle estimates $PS(c, y)$ for all $c \in \mathcal{C}$, Alg.~\ref{alg:main} is sound and complete.
\end{theorem*}

\emph{Proof.} Soundness and completeness follow directly from the specification of (P1) $\mathcal{C}$ and (P2) $\preceq$ in the algorithm's input $\mathcal{B}$, along with (P3) access to oracle estimates $PS(c, y)$ for all $c \in \mathcal{C}$. Recall that the partial ordering must be complete and transitive, as noted in 
Sect.~\ref{sec:framework}.

Assume that Alg.~\ref{alg:main} generates a false positive, i.e. outputs some $c$ that is not $\tau$-minimal. Then by Def.~\ref{def:minimality}, either the algorithm failed to properly evaluate $PS(c, y)$, thereby violating (P3); or failed to identify some $c'$ such that (i) $PS(c', y) \geq \tau$ and (ii) $c' \prec c$. (i) is impossible by (P3), and (ii) is impossible by (P2). Thus there can be no false positives.

Assume that Alg.~\ref{alg:main} generates a false negative, i.e. fails to output some $c$ that is in fact $\tau$-minimal. By (P1), this $c$ cannot exist outside the finite set $\mathcal{C}$. Therefore there must be some $c \in \mathcal{C}$ for which either the algorithm failed to properly evaluate $PS(c, y)$, thereby violating (P3); or wrongly identified some $c'$ such that (i) $PS(c', y) \geq \tau$ and (ii) $c' \prec c$. Once again, (i) is impossible by (P3), and (ii) is impossible by (P2). Thus there can be no false negatives.

\subsubsection{Proof of Theorem \ref{thm:ump}}

\begin{theorem*}
    With sample estimates $\hat{PS}(c, y)$ for all $c \in \mathcal{C}$, Alg.~\ref{alg:main} is uniformly most powerful.
\end{theorem*}

\emph{Proof.} A testing procedure is uniformly most powerful (UMP) if it attains the lowest type II error $\beta$ of all tests with fixed type I error $\alpha$. Let $\Theta_0, \Theta_1$ denote a partition of the parameter space into null and alternative regions, respectively. The goal in frequentist inference is to test the null hypothesis $H_0: \theta \in \Theta_0$ against the alternative $H_1: \theta \in \Theta_1$ for some parameter $\theta$. Let $\psi(X)$ be a testing procedure of the form $\mathds{1}[T(X) \geq c_{\alpha}]$, where $X$ is a finite sample, $T(X)$ is a test statistic, and $c_{\alpha}$ is the critical value. This latter parameter defines a rejection region such that test statistics integrate to $\alpha$ under $H_0$. We say that $\psi(X)$ is UMP iff, for any other test $\psi'(X)$ such that 
\begin{align*}
    \sup_{\theta \in \Theta_0} \mathds{E}_{\theta}[\psi'(X)] \leq \alpha,
\end{align*}
we have
\begin{align*}
    (\forall \theta \in \Theta_1) ~\mathds{E}_{\theta}[\psi'(X)] \leq \mathds{E}_{\theta}[\psi(X)],
\end{align*}
where $\mathds{E}_{\theta \in \Theta_1}[\psi(X)]$ denotes the power of the test to detect the true $\theta$, $1 - \beta_{\psi}(\theta)$. The UMP-optimality of Alg.~\ref{alg:main} follows from the UMP-optimality of the binomial test (see \citep[Ch.~3]{Lehmann2005}), which is used to decide between $H_0: PS(c, y) < \tau$ and $H_1: PS(c, y) \geq \tau$ on the basis of observed proportions $\hat{PS}(c, y)$, estimated from $n$ samples for all $c \in \mathcal{C}$. The proof now takes the same structure as that of 
Thm.~\ref{thm:complete}, with (P3) replaced by (P$3'$): access to UMP estimates of $PS(c, y)$. False positives are no longer impossible but bounded at level $\alpha$; false negatives are no longer impossible but occur with frequency $\beta$. Because no procedure can find more $\tau$-minimal factors for any fixed $\alpha$, 
Alg.~\ref{alg:main} is UMP.

\subsection{Propositions}

\subsubsection{Proof of Proposition \ref{prop:shap1}
}
\begin{prop*} 
    Let $c_S(\bm{z}) = 1$ iff $\bm{x} \subseteq \bm{z}$ was constructed by holding $\bm{x}^S$ fixed and sampling $\bm{X}^R$ according to $\mathcal{D}(\cdot|S)$. Then $v(S) = PS(c_S, y)$.
\end{prop*}
As noted in the text, $\mathcal{D}(\bm{x}|S)$ may be defined in a variety of ways (e.g., via marginal, conditional, or interventional distributions). For any given choice, let $c_S(\bm{z}) = 1$ iff $\bm{x}$ is constructed by holding $\bm{x}^S_i$ fixed and sampling $\bm{X}^R$ according to $\mathcal{D}(\bm{x}|S)$. Since we assume binary $Y$ (or binarized, as discussed in Sect.~\ref{sec:framework}), we can rewrite Eq.~\ref{eq:val} as a probability:
\begin{align*}
    v(S) = P_{\mathcal{D}(\bm{x}|S)} (f(\bm{x}_i) = f(\bm{x})),
\end{align*}
where $\bm{x}_i$ denotes the input point. Since conditional sampling is equivalent to conditioning after sampling, this value function is equivalent to $PS(c_S, y)$ by Def. \ref{def:probsuff}.

\subsubsection{Proof of Proposition \ref{prop:anchors}}
\begin{prop*} 
    Let $c_A(\bm{z}) = 1$ iff $A(\bm{x}) = 1$. Then $\text{prec}(A) = PS(c_A, y)$.
\end{prop*}
The proof for this proposition is essentially identical, except in this case our conditioning event is $A(\bm{x}) = 1$. Let $c_A = 1$ iff $A(\bm{x}) = 1$. Precision prec($A$), given by the lhs of Eq. \ref{eq:shapley}, is defined over a conditional distribution $\mathcal{D}(\bm{x}|A)$. Since conditional sampling is equivalent to conditioning after sampling, this probability reduces to $PS(c_A, y)$.

\subsubsection{Proof of Proposition \ref{prop:cf}}
\begin{prop*} 
    Let $cost$ be a function representing $\preceq$, and let $c$ be some factor spanning reference values. Then the counterfactual recourse objective is:
    \begin{equation}
        c^* = \argmin_{c \in \mathcal{C}} ~cost(c) ~~\textrm{s.t.} ~PS(c, 1 - y) \geq \tau,
    \end{equation}
    where $\tau$ denotes a decision threshold. Counterfactual outputs will then be any $\bm{z} \sim \mathcal{D}$ such that $c^*(\bm{z}) = 1$. 
\end{prop*}
There are two closely related ways of expressing the counterfactual objective: as a search for optimal \emph{points}, or optimal \emph{actions}. We start with the latter interpretation, reframing actions as factors. We are only interested in solutions that flip the original outcome, and so we constrain the search to factors that meet an I2R sufficiency threshold, $PS(c, 1-y) \geq \tau$. Then the optimal action is attained by whatever factor (i) meets the sufficiency criterion and (ii) minimizes cost. Call this factor $c^*$. The optimal point is then any $\bm{z}$ such that $c^*(\bm{z}) = 1$. 

\subsubsection{Proof of Proposition \ref{prop:pearl}}
\begin{prop*}
    Consider the bivariate Boolean setting, as in Sect.~\ref{sec:nec_suf}. We have two counterfactual distributions: an input space $\mathcal{I}$, in which we observe $x, y$ but intervene to set $X = x'$; and a reference space $\mathcal{R}$, in which we observe $x', y'$ but intervene to set $X = x$. Let $\mathcal{D}$ denote a uniform mixture over both spaces, and let auxiliary variable $W$ tag each sample with a label indicating whether it comes from the original ($W = 1$) or contrastive ($W = 0$) counterfactual space. Define $c(\bm{z}) = w$. Then we have $\texttt{suf}(x, y) = PS(c, y)$ and $\texttt{nec}(x, y) = PS(1 - c, y')$.
\end{prop*}
Recall from Sect. \ref{sec:nec_suf} that \citet[Ch.~9]{Pearl2000} defines $\texttt{suf}(x, y) := P(y_x|x', y')$ and $\texttt{nec}(x, y) := P(y'_{x'}|x,y).$ We may rewrite the former as $P_{\mathcal{R}}(y)$, where the reference space $\mathcal{R}$ denotes a counterfactual distribution conditioned on $x', y', do(x)$. Similarly, we may rewrite the latter as $P_{\mathcal{I}}(y')$, where the input space $\mathcal{I}$ denotes a counterfactual distribution conditioned on $x, y, do(x')$. Our context $\mathcal{D}$ is a uniform mixture over both spaces.

The key point here is that the auxiliary variable $W$ indicates whether samples are drawn from $\mathcal{I}$ or $\mathcal{R}$. Thus conditioning on different values of $W$ allows us to toggle between probabilities over the two spaces. Therefore, for $c(\bm{z}) = w$, we have $\texttt{suf}(x, y) = PS(c, y)$ and $\texttt{nec}(x, y) = PS(1 - c, y')$.

\section{Additional discussions of method}\label{sec:app_method}

\subsection{$\tau$-minimality and necessity}
\begin{figure}[!ht]
    \centering
    \includegraphics[scale=0.3]{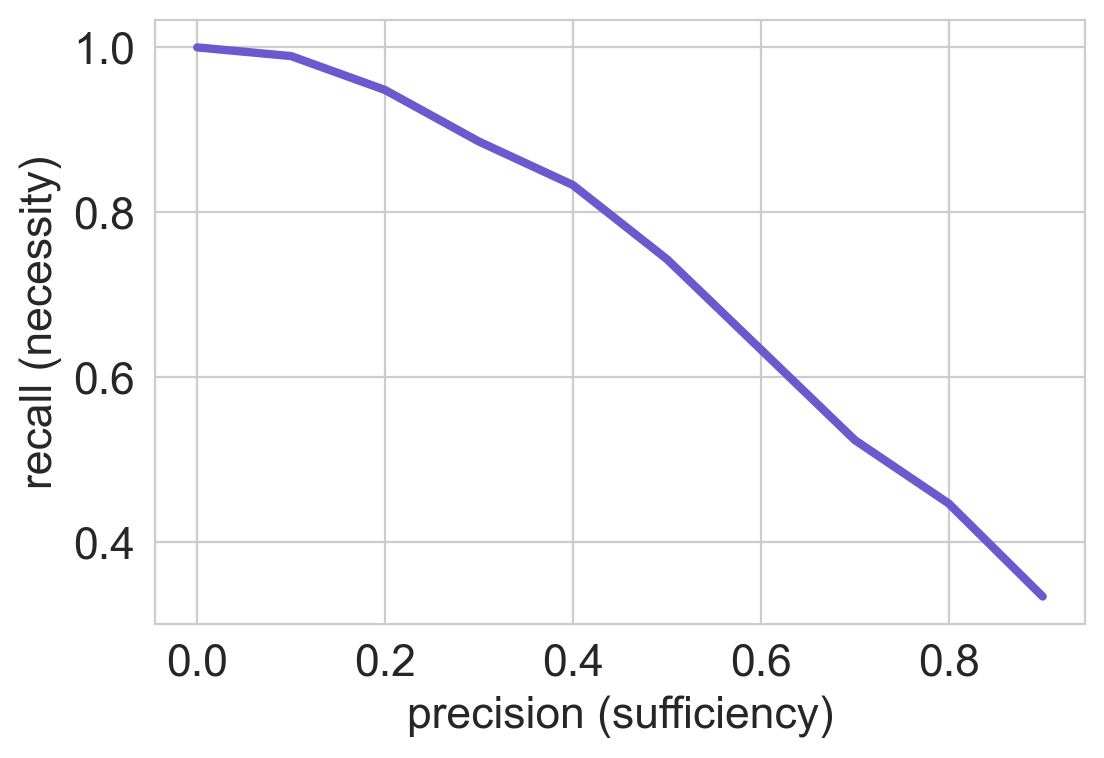}
    \caption{An example curve exemplifying the relationship between $\tau$ and cumulative probability necessity attained by selected $\tau$-minimal factors.}
    \label{fig:prec_recall_supp}
\end{figure}

As a follow up to Remark \ref{rmk:tau} in Sect. \ref{sec:lens}, we expand here upon the relationship between $\tau$ and cumulative probabilities of necessity, which is similar to a precision-recall curve quantifying and qualifying errors in classification tasks. In this case, as we lower $\tau$, we allow more factors to be taken into account, thus covering more pathways towards a desired outcome in a cumulative sense. We provide an example of such a precision-recall curve in Fig. \ref{fig:prec_recall_supp}, using an R2I view of the \texttt{German} credit dataset. Different levels of cumulative necessity may be warranted for different tasks, depending on how important it is to survey multiple paths towards an outcome. Users can therefore adjust $\tau$ to accommodate desired levels of cumulative $PN$ over successive calls to LENS.

\section{Additional discussions of experimental results}\label{sec:app_exp}

\subsection{Data pre-processing and model training}
\textbf{German Credit Risk.} We first download the dataset from Kaggle,\footnote{See \url{https://www.kaggle.com/kabure/german-credit-data-with-risk?select=german_credit_data.csv}.} which is a slight modification of the UCI version \citep{UCI_Dua:2019}. We follow the pre-processing steps from a Kaggle tutorial.\footnote{See \url{https://www.kaggle.com/vigneshj6/german-credit-data-analysis-python}.} In particular, we map the categorical string variables in the dataset (\texttt{Savings}, \texttt{Checking}, \texttt{Sex}, \texttt{Housing}, \texttt{Purpose} and the outcome \texttt{Risk}) to numeric encodings, and mean-impute values missing values for \texttt{Savings} and \texttt{Checking}. We then train an Extra-Tree classifier \citep{Geurts2006} using scikit-learn, with random state 0 and max depth 15. All other hyperparameters are left to their default values. The model achieves a 71\% accuracy.

\textbf{German Credit Risk - Causal.} We assume a partial ordering over the features in the dataset, as described in Fig. \ref{fig:recourse_DAG}. We use this DAG to fit a structural causal model (SCM) based on the original data. In particular, we fit linear regressions for every continuous variable and a random forest classifier for every categorical variable. When sampling from $\mathcal{D}$, we let variables remain at their original values unless either (a) they are directly intervened on, or (b) one of their ancestors was intervened on. In the latter case, changes are propagated via the structural equations. We add stochasticity via Gaussian noise for continuous outcomes, with variance given by each model's residual mean squared error. For categorical variables, we perform multinomial sampling over predicted class probabilities. We use the same $f$ model as for the non-causal \texttt{German} credit risk description above.

\vspace{-1ex}
\textbf{SpamAssassins.} The original spam assassins dataset comes in the form of raw, multi-sentence emails captured on the Apache SpamAssassins project, 2003-2015.\footnote{\raggedright See \url{https://spamassassin.apache.org/old/credits.html}.} We segmented the emails to the following ``features'': \texttt{From} is the sender; \texttt{To} is the recipient; \texttt{Subject} is the email's subject line; \texttt{Urls} records any URLs found in the body; \texttt{Emails} denotes any email addresses found in the body; \texttt{First Sentence}, \texttt{Second Sentence}, \texttt{Penult Sentence}, and \texttt{Last Sentence} refer to the first, second, penultimate, and final sentences of the email, respectively. We use the original outcome label from the dataset (indicated by which folder the different emails were saved to). Once we obtain a dataset in the form above, we continue to pre-process by lower-casing all characters, only keeping words or digits, clearing most punctuation (except for `-' and `\_'), and removing stopwords based on nltk's provided list \citep{bird2009natural}. Finally, we convert all clean strings to their mean 50-dim GloVe vector representation \citep{pennington2014glove}. We train a standard MLP classifier using scikit-learn, with random state 1, max iteration 300, and all other hyperparameters set to their default values.\footnote{\raggedright See \url{https://scikit-learn.org/stable/modules/generated/sklearn.\\neural\_network.MLPClassifier.html}.} This model attains an accuracy of 98.3\%.

\textbf{IMDB.} We follow the pre-processing and modeling steps taken in a standard tutorial on LSTM training for sentiment prediction with the IMDB dataset.\footnote{\raggedright See \url{https://github.com/hansmichaels/sentiment-analysis-IMDB-Review-using-LSTM/blob/master/sentiment_analysis.py.ipynb}.} The CSV is included in the repository named above, and can be additionally downloaded from Kaggle or ai.standford.\footnote{\raggedright See \url{https://www.kaggle.com/lakshmi25npathi/imdb-dataset-of-50k-movie-reviews} or \url{http://ai.stanford.edu/\~amaas/data/sentiment/}.} In particular, these include removal of HTML-tags, non-alphabetical characters, and stopwords based on the the list provided in the ntlk package, as well as changing all alphabetical characters to lower-case. We then train a standard LSTM model, with 32 as the embedding dimension and 64 as the dimensionality of the output space of the LSTM layer, and an additional dense layer with output size 1. We use the sigmoid activation function, binary cross-entropy loss, and optimize with Adam \citep{kingma_adam}. All other hyperparameters are set to their default values as specified by Keras.\footnote{\raggedright See \url{https://keras.io}.} The model achieves an accuracy of 87.03\%.

\begin{table*}[t!]
\caption{Recourse options for a single input given by DiCE and our method. We report targets of interventions as suggested options, but they could correspond to different values of interventions. Our method tends to propose more minimal and diverse intervention targets. Note that all of DiCE's outputs are already subsets of LENS's two top suggestions, and due to $\tau$-minimality LENS is forced to pick the next factors to be non-supersets of the two top rows. This explains the higher cost of LENS's bottom three rows.}
\vspace{-2ex}
\centering
\resizebox{\textwidth}{!}{%
\begin{tabular}{|llllllll|llll|}
\toprule
\rowcolor[HTML]{FE996B} 
\multicolumn{8}{|l|}{\cellcolor[HTML]{FE996B}input} &
  \multicolumn{2}{l|}{\cellcolor[HTML]{C0C0C0}DiCE output} &
  \multicolumn{2}{l|}{\cellcolor[HTML]{C0C0C0}\textbf{LENS output}} \\ \midrule
\rowcolor[HTML]{EFEFEF} 
\multicolumn{1}{|l|}{\cellcolor[HTML]{EFEFEF}Age} &
  \multicolumn{1}{l|}{\cellcolor[HTML]{EFEFEF}Wrkcls} &
  \multicolumn{1}{l|}{\cellcolor[HTML]{EFEFEF}Edu.} &
  \multicolumn{1}{l|}{\cellcolor[HTML]{EFEFEF}Marital} &
  \multicolumn{1}{l|}{\cellcolor[HTML]{EFEFEF}Occp.} &
  \multicolumn{1}{l|}{\cellcolor[HTML]{EFEFEF}Race} &
  \multicolumn{1}{l|}{\cellcolor[HTML]{EFEFEF}Sex} &
  Hrs/week &
  \multicolumn{1}{l|}{\cellcolor[HTML]{EFEFEF}Targets of intervention} &
  \multicolumn{1}{l|}{\cellcolor[HTML]{EFEFEF}Cost} &
  \multicolumn{1}{l|}{\cellcolor[HTML]{EFEFEF}Targets of intervention} &
  Cost \\ \midrule
 &
   &
   &
   &
   &
   &
   &
   &
  Age, Edu., Marital, Hrs/week &
  8.13 &
  Edu. &
  1 \\
 &
   &
   &
   &
   &
   &
   &
   &
  Age, Edu., Marital, Occp., Sex, Hrs/week &
  5.866 &
  Martial &
  1 \\
 &
   &
   &
   &
   &
   &
   &
   &
  Age, Wrkcls, Educ., Marital, Hrs/week &
  5.36 &
  Occp., Hrs/week &
  19.3 \\
 &
   &
   &
   &
   &
   &
   &
   &
  Age, Edu., Occp., Hrs/week &
  3.2 &
  Wrkcls, Occp., Hrs/week &
  12.6 \\
\multirow{-5}{*}{42} &
  \multirow{-5}{*}{Govt.} &
  \multirow{-5}{*}{HS-grad} &
  \multirow{-5}{*}{Single} &
  \multirow{-5}{*}{Service} &
  \multirow{-5}{*}{White} &
  \multirow{-5}{*}{Male} &
  \multirow{-5}{*}{40} &
  Edu., Hrs/week &
  11.6 &
  Age, Wrkcls, Occp., Hrs/week &
  12.2 \\ \bottomrule
\end{tabular}%
}
\label{tab:Dice_single_example}
\end{table*}

\textbf{Adult Income.} We obtain the adult income dataset via DiCE's implementation\footnote{See \url{https://github.com/interpretml/DiCE}.} and followed Haojun Zhu's pre-processing steps.\footnote{See \url{https://rpubs.com/H_Zhu/235617}.} For our recourse comparison, we use a pretrained MLP model provided by the authors of DiCE, which is a single layer, non-linear model trained with TensorFlow and stored in their repository as `adult.h5'.

\subsection{Tasks}

\begin{figure}[!hbtp]
    \centering
    \includegraphics[scale=0.45]{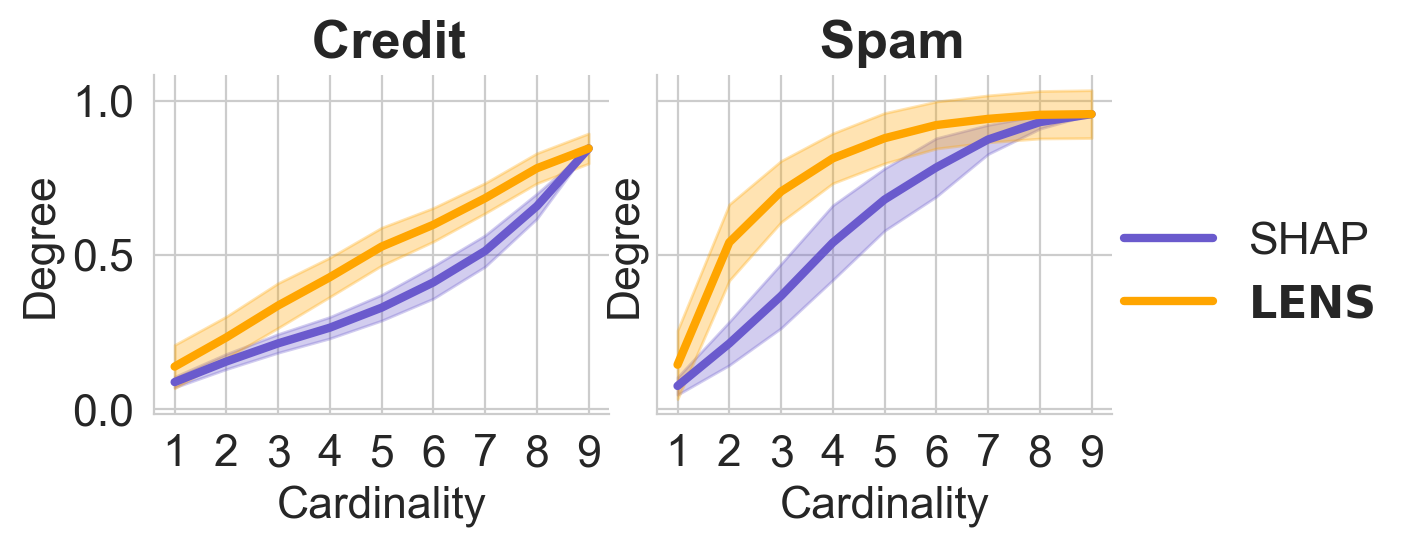}
    \caption{Comparison of degrees of sufficiency in I2R setting, 
    for top $k$ features based on SHAP scores, against the best performing subset of cardinality $k$ identified by our method. Results for \texttt{German} are averaged over 50 inputs; results for \texttt{SpamAssassins} are averaged over 25 inputs.}
    \label{fig:credit_spam_shap_i2r_supp}
\end{figure}

\textbf{Comparison with attributions.} For completeness, we also include here comparison of cumulative attribution scores per cardinality with probabilities of sufficiency for the I2R view (see Fig. \ref{fig:credit_spam_shap_i2r_supp}).

\textbf{Sentiment sensitivity analysis.} We identify sentences in the original \texttt{IMDB} dataset that are up to 10 words long. Out of those, for the first example we only look at wrongly predicted sentences to identify a suitable example. For the other example, we simply consider a random example from the 10-word maximum length examples. We noted that Anchors uses stochastic word-level perturbations for this setting. This leads them to identify explanations of higher cardinality for some sentences, which include elements that are not strictly necessary. In other words, their outputs are not minimal, as required for descriptions of ``actual causes'' \citep{halpern2005causes1, halpern2016actual}.

\textbf{Comparison with Anchors.} To complete the picture of our comparison with Anchors on the \texttt{German} Credit Risk dataset, we provide here additional results. In the main text, we included a comparison of Anchors's single output precision against the mean degree of sufficiency attained by our multiple suggestions per input. We sample 100 different inputs from the \texttt{German} Credit dataset and repeat this same comparison. Here we additionally consider the minimum and maximum $PS(c, y)$ attained by LENS against Anchors. Note that even when considering minimum $PS$ suggestions by LENS, i.e. our worst output, the method shows more consistent performance. We qualify this discussion by noting that Anchors may generate results comparable to our own by setting the $\delta$ hyperparameter to a lower value. However, \citet{Ribeiro2018} do not discuss this parameter in detail in either their original article or subsequent notebook guides. They use default settings in their own experiments, and we expect most practitioners will do the same. 

\begin{figure}[ht]
    \centering
    \subfloat{\includegraphics[scale=0.25]{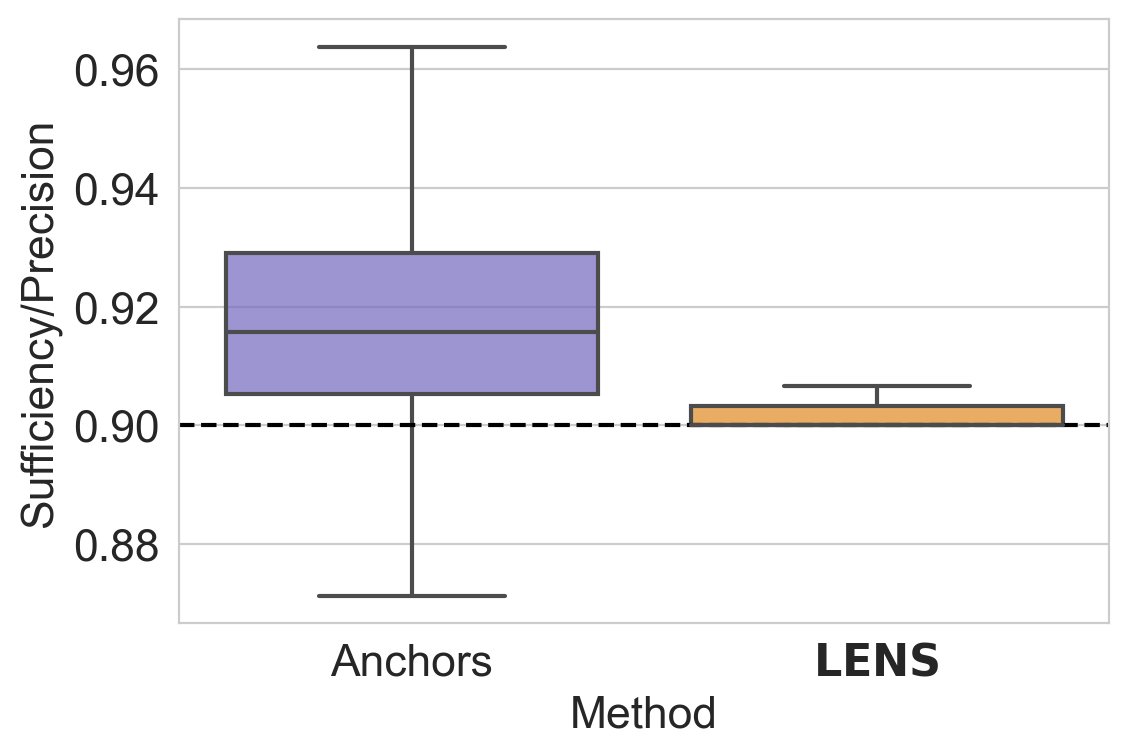}}
    \qquad
    \subfloat{\includegraphics[scale=0.25]{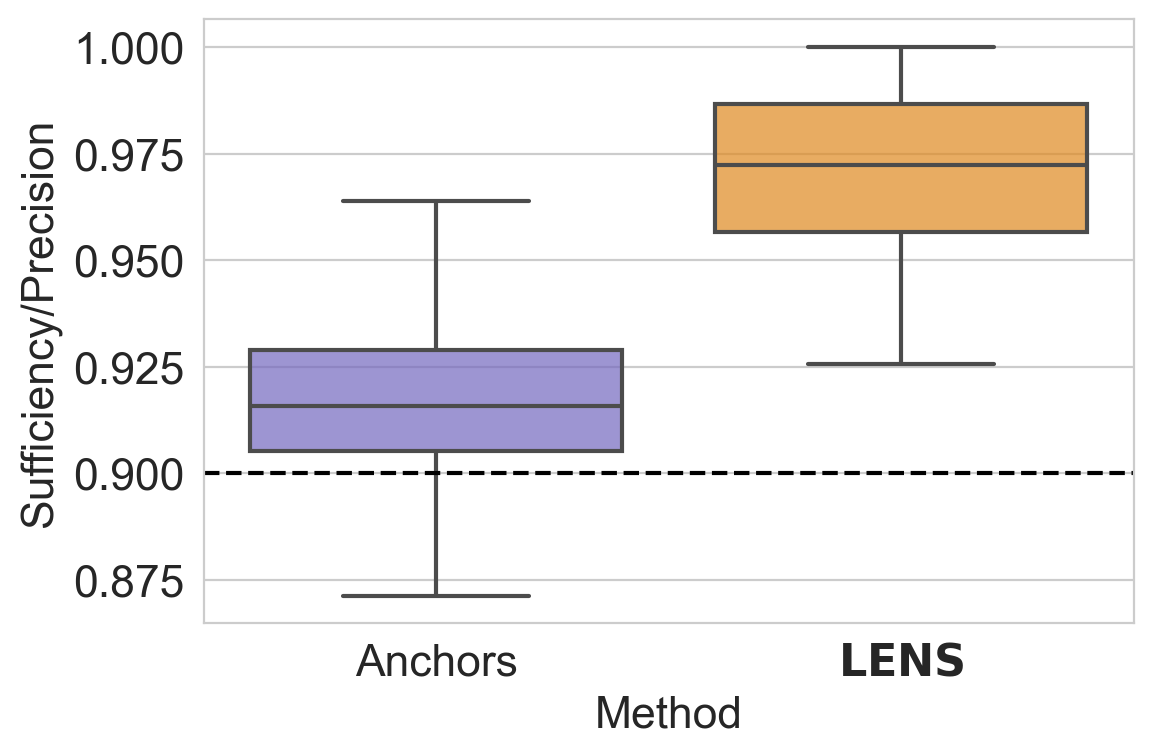}}
  \hfill
  \caption{We compare degree of sufficiency against precision scores attained by the output of LENS and Anchors for examples from \texttt{German}. We repeat the experiment for 100 sampled inputs, and each time consider the single output by Anchors against the min (left) and max (right) $PS(c, y)$ among LENS's multiple candidates. Dotted line indicates $\tau=0.9$, the threshold we chose for this experiment.}
    \label{fig:Dice_comp2}%
\end{figure}

\textbf{Recourse: DiCE comparison}
First, we provide a single illustrative example of the lack of diversity in intervention targets we identify in DiCE's output. Let us consider one example, shown in Table \ref{tab:Dice_single_example}. While DiCE outputs are diverse in terms of values and target combinations, they tend to have great overlap in intervention targets. For instance, \texttt{Age} and \texttt{Education} appear in almost all of them. Our method would focus on minimal paths to recourse that would involve different combinations of features.

\begin{figure}[!hbtp]
    \centering
    
    \subfloat{\includegraphics[scale=0.25]{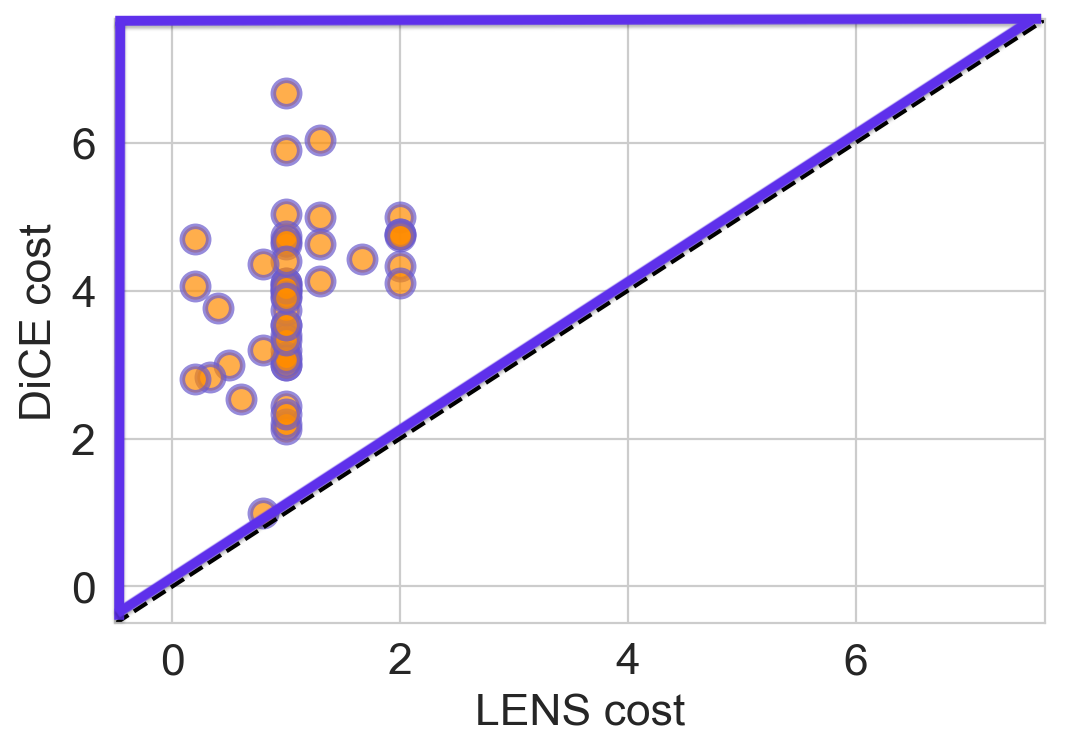}}
    \qquad
    \subfloat{\includegraphics[scale=0.25]{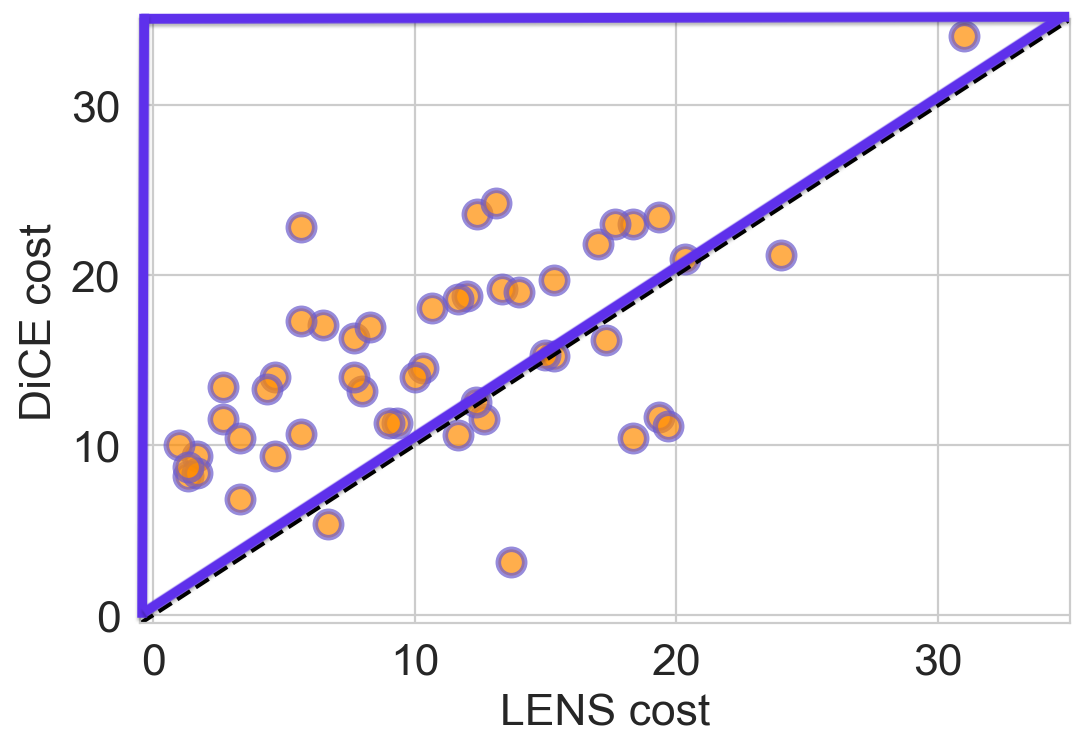}}
  \hfill
  \caption{We show results over 50 input points sampled from the original dataset, and all possible references of the opposite class, across two metrics: the min cost (left) of counterfactuals suggested by our method vs. DiCE, and the max cost (right) of counterfactuals.}
    \label{fig:Dice_comp}%
\end{figure}

Next, we also provide additional results from our cost comparison with DiCE's output in Fig.~\ref{fig:Dice_comp2}. While in the main text we include a comparison of our mean cost output against DiCE's, here we additionally include a comparison of min and max cost of the methods' respective outputs. We see that even when considering minimum and maximum cost, our method tends to suggest lower cost recourse options. In particular, note that all of DiCE's outputs are already subsets of LENS's two top suggestions. The higher costs incurred by LENS for the next two lines are a reflection of this fact: due to $\tau$-minimality, LENS is forced to find other interventions that are no longer supersets of options already listed above.



\end{document}